\documentclass[default,iicol]{sn-jnl}

\usepackage{times}

\usepackage{graphicx}
\usepackage{amsmath,amssymb,amsfonts}%
\usepackage{amsthm}%
\usepackage{mathrsfs}%
\usepackage[title]{appendix}%
\usepackage[dvipsnames,table,xcdraw]{xcolor}
\usepackage{textcomp}%
\usepackage{manyfoot}%
\usepackage{booktabs}%
\usepackage{algorithm}%
\usepackage{algorithmicx}%
\usepackage{algpseudocode}%
\usepackage{listings}%
\usepackage{xspace}%
\usepackage{multirow}%
\usepackage{caption}
\usepackage{subcaption}
\usepackage{wrapfig}
\usepackage{makecell} 
\usepackage{pifont}
\newcommand{\cmark}{\ding{51}}%
\usepackage{svg}
\usepackage{pgfplots}
\pgfplotsset{compat=1.10}
\usepackage{tikz}
\usetikzlibrary{fit,matrix,positioning, 
decorations.pathreplacing,calc,shapes,arrows,shadows,patterns}
\usepackage[normalem]{ulem}

\usepackage[symbol]{footmisc}

\usepackage{trimclip} 
\usepackage[export]{adjustbox}

\newcommand{\gilles}[1] {\textcolor{black}{#1}} 
\newcommand{\OS}[1] {\textcolor{black}{#1}} 
\newcommand{\PP}[1] {\textcolor{black}{#1}}

\newcommand{\red}[1]{{\color{red}{#1}}}

\newcommand{\citeme}[1]{\red{[XX]}}
\newcommand{\refme}[1]{\red{(XX)}}

\newcommand{\tran}{^\top}

\newcommand{\real}{\mathbb{R}}

\newcommand{\abs}[1]{\left|{#1}\right|}

\newcommand{\vA}{\mathbf{A}}

\newcommand{\vD}{\mathbf{D}}

\newcommand{\vI}{\mathbf{I}}

\newcommand{\vK}{\mathbf{K}}

\newcommand{\vQ}{\mathbf{Q}}

\newcommand{\vV}{\mathbf{V}}

\newcommand{\vX}{\mathbf{X}}
\newcommand{\vY}{\mathbf{Y}}

\newcommand{\vf}{\mathbf{f}}

\newcommand{\vm}{\mathbf{m}}

\newcommand{\Vs}{\mathbf{s}}

\makeatletter
\newcommand*\bdot{\mathpalette\bdot@{.7}}
\newcommand*\bdot@[2]{\mathbin{\vcenter{\hbox{\scalebox{#2}{$\m@th#1\bullet$}}}}}
\makeatother

\makeatletter
\DeclareRobustCommand\onedot{\futurelet\@let@token\@onedot}
\def\@onedot{\ifx\@let@token.\else.\null\fi\xspace}

\makeatother

\def \cls{\texttt{CLS}\xspace}  

\newcommand{\edge}{\vA}

\newcommand{\indicatory}{\mathbf{y}}

\newcommand{\rn}[1] {\textcolor{PineGreen!90!black}{#1}}

\newcommand{\improv}[1] {\bf{\textcolor{Green}{+#1}}}

\newcommand{\unsupsaliency}{unsupervised saliency detection}
\newcommand{\singleobjectdiscovery}{single-object discovery}
\newcommand{\multiclassif}{class-agnostic multi-object detection}
\newcommand{\multiseg}{class-agnostic instance segmentation}

\newcommand{\Unsupsaliency}{Unsupervised saliency detection}
\newcommand{\Singleobjectdiscovery}{Single-object discovery}
\newcommand{\Multiclassif}{Class-agnostic multi-object detection}
\newcommand{\Multiseg}{Class-agnostic instance segmentation}

\newcommand{\tabletitle}[1]{\cellcolor{PineGreen!30!white}{\texttt{#1}}}

\newcommand{\parag}[1]{\vspace{5pt}\noindent\textbf{\emph{#1}}}

\begin{document}

\pgfkeys{/pgf/number format/.cd,1000 sep={\,}}

\title[]{How far can we go without manual annotation? A survey of unsupervised object localization methods}
\title[]{Unsupervised Object Localization in the Era of Self-Supervised ViTs: A Survey}

\author{\fnm{Oriane} \sur{Siméoni}\textsuperscript{1}}
\author{\fnm{Éloi} \sur{Zablocki}\textsuperscript{1}}
\author{\fnm{Spyros} \sur{Gidaris}\textsuperscript{1}}
\author{\fnm{Gilles} \sur{Puy}\textsuperscript{1}}
\author{\fnm{Patrick} \sur{Pérez}\textsuperscript{2}$^{	\dagger}$}
\affil{\textsuperscript{1}valeo.ai, Paris, France \hspace{5mm} \textsuperscript{2}Kyutai, Paris, France
\vspace{1mm}
}

\abstract{The recent enthusiasm for \emph{open-world} vision systems shows the high interest of the community to perform perception tasks outside of the closed-vocabulary benchmark setups which have been so popular until now. Being able to discover objects in images and videos without knowing in advance what objects populate the dataset is an exciting prospect. \emph{But how to find objects without knowing anything about them?} 
Recent works show that it is possible to perform class-agnostic \emph{unsupervised object localization} by exploiting \emph{self-supervised pre-trained features}. 
We propose here a survey of \emph{unsupervised object localization} methods that discover objects in images \emph{without requiring any manual annotation} in the era of self-supervised ViTs.
}


\maketitle
\footnotetext[2]{Work done at valeo.ai.}

\section{Introduction}
\label{introduction}

Object localization in 2D images is a key task for many perception systems, e.g., autonomous robots and cars, augmented reality headsets, visual search engines, etc. Depending on the application, the task can take different forms such as object detection \cite{ren2015fasterrcnn,carion2020detr} and instance segmentation \cite{he2017mask, chen2018deeplab}.
Achieving good results in these tasks has traditionally hinged on one critical factor: access to extensive, meticulously annotated datasets \citep{coco2014,pascal-voc-2007} to build and train deep neural networks.

However, this paradigm comes with inherent limitations.
The first obvious one is the high \emph{cost} and \emph{tediousness} involved in acquiring these datasets. The second limitation stems from the \emph{finite} and \emph{pre-defined} nature of the set of object classes which significantly narrows the scope of what supervised models can perceive and identify.
This becomes particularly problematic in contexts where the ability to recognize and respond to unknown or unconventional objects is crucial.
For example, in autonomous driving applications, where the road can present a multitude of unpredictable elements, the need to apprehend whatever comes into view is paramount.

The high interest around the recent Segment Anything (SAM) \cite{kirillov2023segment} model shows the desire for the community to segment any object in an image in a class-agnostic fashion. Although the results obtained with the fully-supervised SAM model 
--- trained on 11M images carefully \emph{annotated} by humans with 1B masks --- are exciting, 
recent works have shown that impressive object localization can be obtained \emph{without} human in the loop. By avoiding to rely on human-made annotation, we can hope to obtain systems which (1) would not suffer human biases, (2) could generalize to new objects / domains.
Moreover, SAM necessitates prompts like points, boxes, or coarse masks to indicate the object to be segmented. 
As opposed to SAM, unsupervised object localization methods are able to find the objects of interest without prompts.

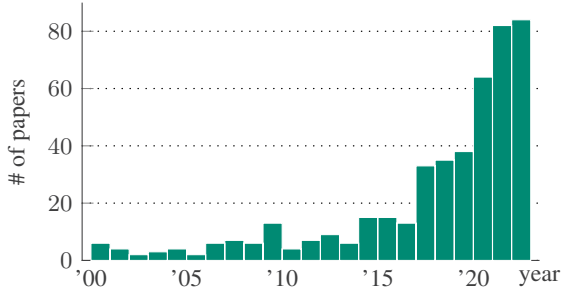
\begin{figure}[ht!]
    \centering
\pgfplotsset{
  grid style = {
    dash pattern = on 0.05mm off 1mm,
    line cap = round,
    black,
    line width = 0.5pt
  }
}
\begin{tikzpicture}
\begin{axis}[
    height=5cm,
    width=\linewidth,
    ymin=0, ymax=90,
    xmax=2023,
    x axis line style=-, 
    bar width=1pt,
    font=\small,
    major x tick style = transparent,
    enlarge x limits=0.02,
    ymajorgrids = true,
    ylabel= \# of papers,
    xlabel = year,
    xlabel shift = 5pt,
    White!10!Black,
    axis x line*=bottom,
    axis y line*=left,
    xtick distance = 5,
    ylabel near ticks,
    xlabel near ticks,
    xlabel style={at={(1,0)},below},
    xtick={2000,2005,2010,2015,2020},
    xticklabels={'00, '05, '10, '15, '20},
    ], 
\addplot+[ybar interval,mark=no,draw=White,fill=PineGreen] plot coordinates { 
        (2023, 84)
        (2022, 82)
        (2021, 64)
        (2020, 38)
        (2019, 35)
        (2018, 33)
        (2017, 13)
        (2016, 15)
        (2015, 15)
        (2014, 6)
        (2013, 9)
        (2012, 7)
        (2011, 4)
        (2010, 13)
        (2009, 6)
        (2008, 7)
        (2007, 6)
        (2006, 2)
        (2005, 4)
        (2004, 3)
        (2003, 2)
        (2002, 4)
        (2001, 6)
        (2000, 2)
        };
\addplot+[ybar interval,mark=no,draw = black, fill = Gray] plot coordinates {  (2023, 60) };
\end{axis}
\end{tikzpicture}
    \caption{\textbf{Evolution of the number of papers on unsupervised object localization.} Histogram of the number of papers mentioning ``unsupervised object detection/segmentation/localization'' in their title per year, from 2000 to 2023. Data captured by querying \url{dblp.org} paper repository.}
    \label{fig:hist-publi}
\end{figure}

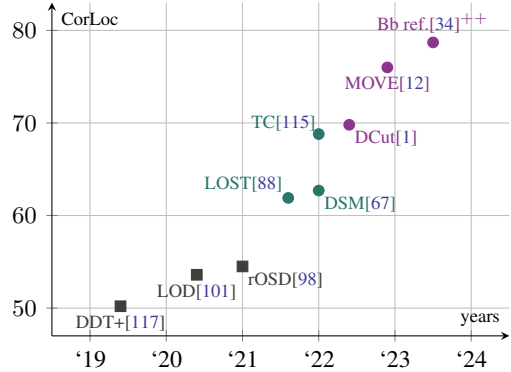
\begin{figure}[t]
    \centering
    \begin{tikzpicture}
    \centering
    \tikzstyle{every node}=[font=\footnotesize]
    \begin{axis}[
        width=\columnwidth,
        height=6cm,
        xmin=2019,
        xmax=2024,
        ymin=50,
        ymax=80,
        xlabel shift=-5 pt,
        ylabel shift=-10 pt,
        font=\footnotesize,
        xlabel=years,
        ylabel=CorLoc,
        label style={font=\small},
        tick label style={font=\small},
        grid=major,
        align =center,
        xtick= {2019, 2020, 2021, 2022, 2023, 2024},    
        xticklabels={`19, `20, `21, `22, `23, `24},
        axis lines = middle,
        enlargelimits = true,
    ]
    \addplot[color=White!10!Black,only marks, mark size=2pt,mark=square*] coordinates {(2019.4,50.2)} node[below, sloped] {DDT+\cite{Wei2019ddtplus}};
    \addplot[color=White!10!Black,only marks, mark size=2pt,mark=square*] coordinates {(2020.4,53.6)} node[below, sloped] {LOD\cite{vo2021largescale}};
    \addplot[color=White!10!Black,only marks, mark size=2pt,mark=square*] coordinates {(2021,54.5)} node[below right=-2pt] {rOSD\cite{Vo20rOSD}};
    
    \addplot[color=PineGreen!70!black,only marks, mark size=2pt] coordinates {(2021.6,61.9)} node[above left=-2pt, sloped] {LOST\cite{simeoni2021lost}};
    \addplot[color=PineGreen!70!black,only marks, mark size=2pt] coordinates {(2022,62.7)} node[below right=-2pt, sloped] {DSM\cite{melas2022deepsectralmethod}};
    \addplot[color=PineGreen!70!black,only marks, mark size=2pt] coordinates {(2022,68.8)} node[above left=-3.5pt, sloped] {TC\cite{wang2022tokencut}};

    \addplot[color=Fuchsia,only marks, mark size=2pt] coordinates {(2022.4,69.8)} node[below right=-2pt, sloped] {DCut\cite{aflalo2022deepcut}};
    \addplot[color=Fuchsia,only marks, mark size=2pt] coordinates {(2022.9,76.0)} node[below, sloped] {MOVE\cite{bielski2022move}};
    \addplot[color=Fuchsia,only marks, mark size=2pt] coordinates {(2023.5,78.7)} node[above, sloped] {Bb ref.\cite{gomel2023boxbasedrefinement}$^{++}$};
    \end{axis}
\end{tikzpicture}
    \caption{\textbf{Performance evolution in unsupervised object localization.} Evolution of the CorLoc score (more details in \autoref{sec:background:def:single-object}) evaluated on VOC07 dataset in the last three years. In \textcolor{Purple}{purple} are methods including a \OS{self-}training stage, 
    when \textcolor{PineGreen!70!black}{green} solely exploit frozen self-supervised features. \OS{\textcolor{White!10!Black}{Gray} squares show previous baselines doing dataset level optimization.} Results have gained more than 20\,pts in 2 years with simpler/faster methods which exploit \emph{self-supervised features}. 
    \OS{`Bb ref.[34]$^{++}$' corresponds to the combination of \cite{gomel2023boxbasedrefinement} with MOVE~\cite{bielski2022move} and the training of a class-agnostic detector following \cite{simeoni2021lost}.}} 
    \label{fig:res-in-time}
\end{figure}

Indeed, a solution to discover objects of interest without relying on annotations is to perform \emph{unsupervised class-agnostic object localization}.
This challenging task consists in \emph{localizing objects} in images with \emph{no human supervision}. Such problem has recently gained a lot of attention as shown by the evolution of the number of papers written about `unsupervised object localization' (see \autoref{fig:hist-publi}).
Moreover, as shown in \autoref{fig:res-in-time}, recent methods (e.g., MOST \cite{rambhatla2023most} and MOVE \cite{bielski2022move}) have achieved impressive results without any annotations, predicting accurate object boxes in over 75\% of the images in the VOC07 \cite{pascal-voc-2007} dataset. 

The recent progress in unsupervised localization tasks owes its success to two critical factors.
First, Vision Transformer (ViT) models \cite{dosovitskiy2021vit} provide global correlations between patches when Convolutional Neural Networks (CNNs) only correlate pixels in a local receptive field.
Second, self-supervised representation learning \cite{caron2021dino, chen2021mocov3, he2022mae} has improved and scaled to massive datasets for feature learning. These techniques can now extract local and global semantically meaningful features from the weak signal provided by pretext tasks.
With such strategies, there is no need to \OS{carefully design hand-crafted methods \cite{yan2013hs, zhu2014wctr}, use generative adversarial models \cite{melas_kyriazi2021_imageseg}, refine noisy labels \cite{nguyen2019deepusps},} nor to interpret the thousands of object proposals generated by handcrafted methods~\cite{zitnick2014edgebox, uijlings2013selectivesearch} (with a high-recall but low precision) using expensive dataset-level quadratic pattern repetition search \cite{Wei2019ddtplus,vo2019unsup_image_matching,vo2020unsup_multi_object_discovery,vo2021largescale}. 

\begin{figure*}
    \centering
    \small
    \begin{subfigure}[t]{0.22\textwidth}
        \centering
        \includegraphics[width=0.9\textwidth]{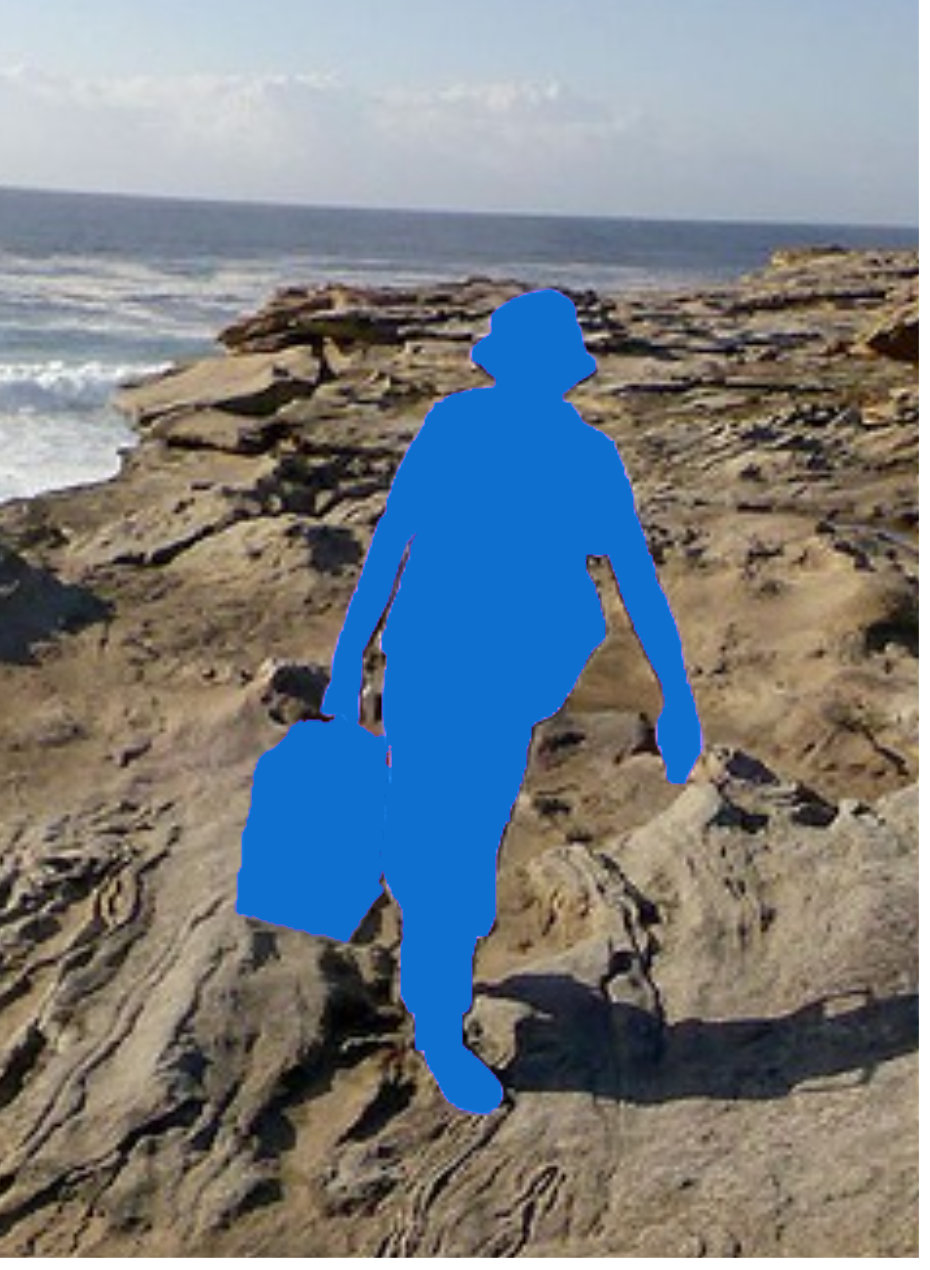}
        \caption{\centering \Unsupsaliency{}}
        \label{fig:foreground-task}
    \end{subfigure}
    \begin{subfigure}[t]{0.22\textwidth}
        \centering
        \includegraphics[width=0.9\textwidth]{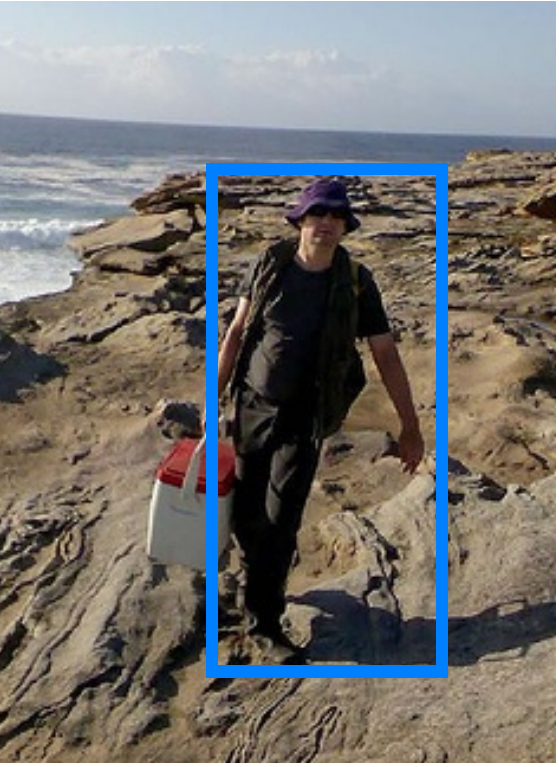}
        \caption{\centering \Singleobjectdiscovery{}}
        \label{fig:single-obj-task}
    \end{subfigure}
    \begin{subfigure}[t]{0.22\textwidth}
        \centering
        \includegraphics[width=0.9\textwidth]{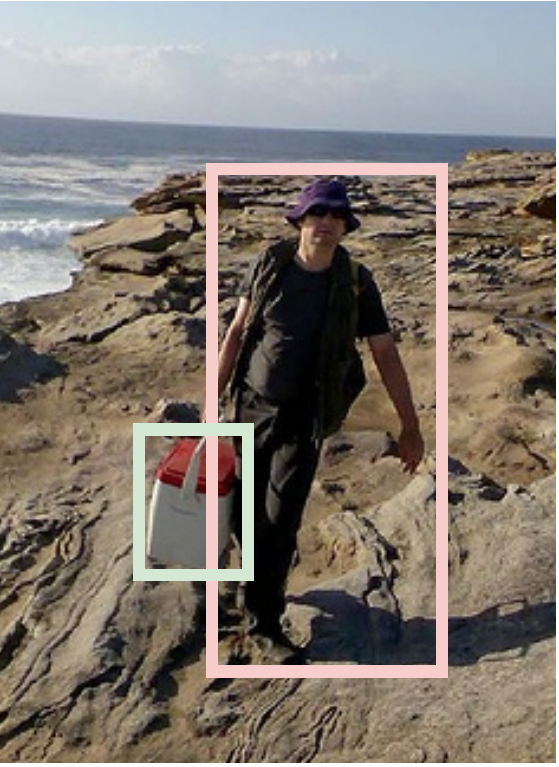}
        \caption{\centering \Multiclassif{}}
        \label{fig:multi-obj-task}
    \end{subfigure}
    \begin{subfigure}[t]{0.22\textwidth}
        \centering
        \includegraphics[width=0.9\textwidth]{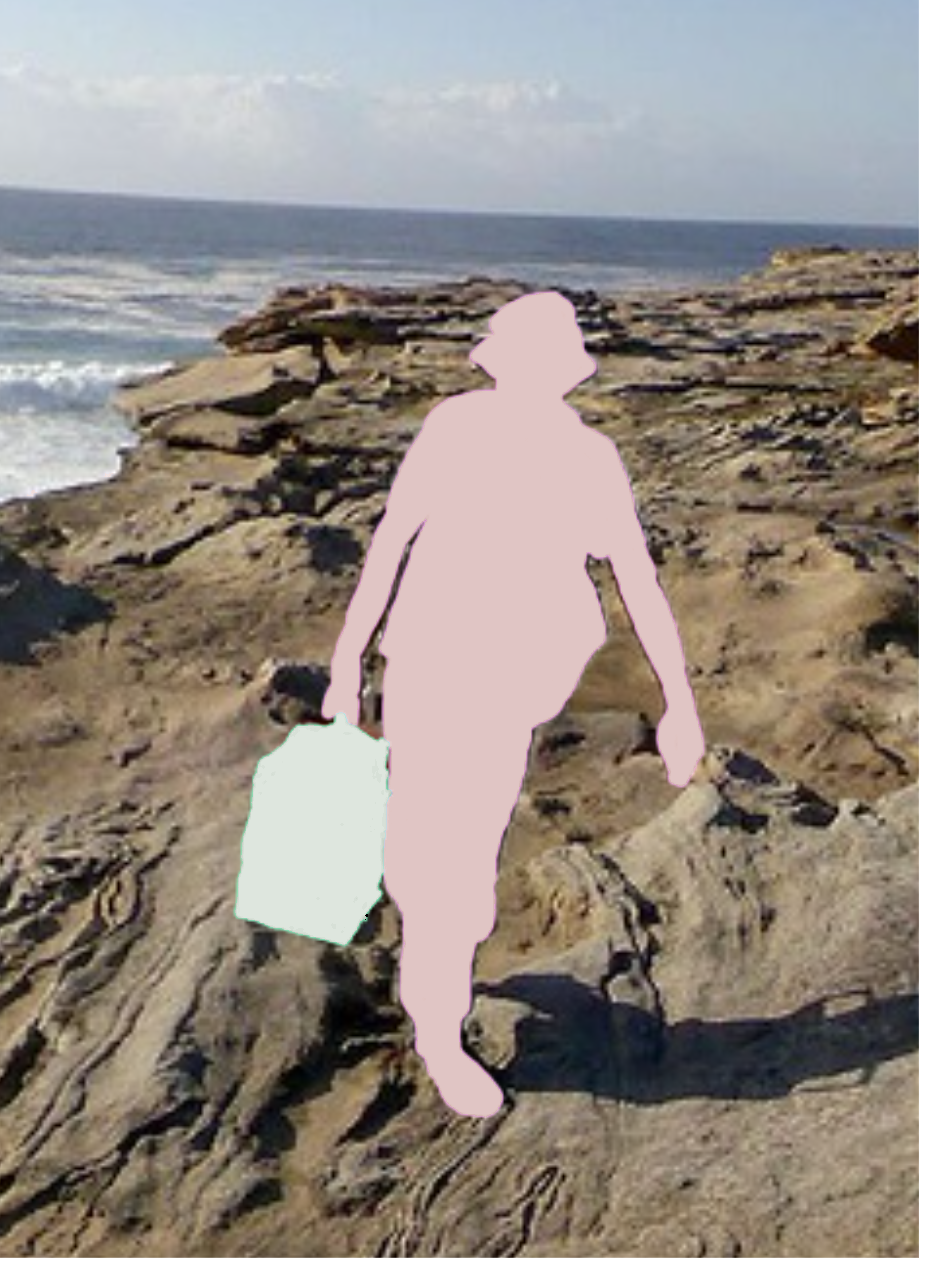}
        \caption{\centering \Multiseg}
        \label{fig:instance-seg}
    \end{subfigure}
    \centering
    \caption{
    \textbf{The different tasks to evaluate \emph{unsupervised object localization} methods}: 
    (a) \emph{\unsupsaliency{}} focuses on foreground/background separation, (b) \emph{\singleobjectdiscovery{}} requires to localize well at least a single object with a box, (c) \emph{\multiclassif{}} evaluates if all objects have been well detected with good boxes and (d) \emph{\multiseg{}} is the analogue with instance masks.}
    \label{fig:unsup-tasks}
\end{figure*}

In this survey, we propose to review \emph{unsupervised object localization methods in the era of self-supervised ViTs}.
We thoroughly present and detail all recent methods addressing this topic and comprehensively organize them for the community. These methods are summarized in \autoref{tab:method_overview}.
To our knowledge, this is the first survey on such topic and we would like to point the reader to related surveys on image classification with limited annotations~\cite{schmarje2021survey_classification}, weakly supervised object localization~\cite{zhang2022weakly_supervised_survey,shao2022weakly_supervised_survey}, object localization on natural scenes \cite{choudhuri2018survey_localization_natural}, object instance segmentation \cite{sharma2022survey_segmentation}, object segmentation \cite{wang2022review_segmentation}, and object detection \cite{amjoud2023review_detection,shehzadi2023detection_transformers}.

The survey is organized as follows \OS{(
state-of-the-art results 
are gathered in \autoref{sec:sota})}:
\begin{itemize}
    \item \autoref{sec:background}: We comprehensively present the different tasks used to evaluate object localization capabilities, as shown in \autoref{fig:unsup-tasks}, as well as the corresponding metrics and typical datasets used to evaluate the methods.
    \item \autoref{sec:ssonly}: We review modern solutions that 
    exploit 
    ViT self-supervised features to produce localization masks in a training-free way.
    \item \autoref{sec:training}: We detail different self-training strategies employed to enhance the quality of coarse localization mined in the self-supervised features. 
    \item \autoref{sec:discussion}: We discuss the success and failures of presented methods and open to different means to obtain object localization information without manual annotation.
\end{itemize}

\begin{table*}
    \centering
    \resizebox{\linewidth}{!}{
    \rowcolors{2}{gray!15}{white}
    \begin{tabular}{@{} l c c l *{4}{c} @{}}
    \toprule
        Method & Sec. & \thead{Code\hspace{-1cm}\\ available\hspace{-1cm}} & \multicolumn{1}{c}{Venue} & \thead{Unsupervised\\ saliency\\ detection} & \thead{Single-object\\ discovery} & \thead{Class-agnostic\\ multi-object\\ detection} & \thead{Class-agnostic\\ multi-object\\ segmentation} \\ 
    \midrule 
        DINO \cite{caron2021dino} & \ref{sec:feats:self} & \cmark & \texttt{ICCV 2021} & --- & \textsuperscript{\phantom{[xx]}}\cmark\textsuperscript{\cite{simeoni2021lost}} & ---  & \textsuperscript{\phantom{[xx]}}\cmark\textsuperscript{\cite{gansbeke2022maskdistill}} \\ 
        LOST \cite{simeoni2021lost} & \ref{sec:no-training:degree}, \ref{sec:training:task-specific} & \cmark & \texttt{BMVC 2021} & \textsuperscript{\phantom{[xx]}}\cmark\textsuperscript{\cite{wang2022tokencut}} & \cmark & \cmark & \textsuperscript{\phantom{[xx]}}\cmark\textsuperscript{\cite{gansbeke2022maskdistill}} \\ 
        SelfMask \cite{shin2022selfmask} & \ref{sec:no-training:clustering}, \ref{sec:training:task-specific} & \cmark & \texttt{CVPRW 2022} & \cmark & \textsuperscript{\phantom{[xx]}}\cmark\textsuperscript{\cite{simeoni2023found}} & --- & --- \\ 
        DSM \cite{melas2022deepsectralmethod} & \ref{sec:no-training:clustering} & \cmark & \texttt{CVPR 2022} & \cmark & \cmark & --- & --- \\ 
        FreeSOLO \cite{wang2022freesolo} & \ref{sec:3:single-to-multi:adaptive}, \ref{sec:training:task-specific} & \cmark & \texttt{CVPR 2022} & \textsuperscript{\phantom{[xx]}}\cmark\textsuperscript{\cite{bielski2022move}} & \textsuperscript{\phantom{[xx]}}\cmark\textsuperscript{\cite{bielski2022move}} & \cmark & \cmark \\ 
        TokenCut \cite{wang2022tokencut} & \ref{sec:no-training:clustering} & \cmark & \texttt{CVPR 2022} & \cmark & \cmark & \textsuperscript{\phantom{[xx, yy]}}\cmark\textsuperscript{\cite{kara2023umod,wang2023cutler}} & \textsuperscript{\phantom{[xx]}}\cmark\textsuperscript{\cite{wang2023cutler}} \\ 
        MOVE \cite{bielski2022move} & \ref{sec:training:head} & \cmark & \texttt{NeurIPS 2022} & \cmark & \cmark & \cmark & --- \\ 
        IMST \cite{lim2022imst} & \ref{sec:no-training}, \ref{sec:training:task-specific}& --- & \texttt{arxiv 2022} & --- & \cmark & \cmark & \cmark \\ 
        MaskDistill \cite{gansbeke2022maskdistill} & \ref{sec:no-training:degree} & --- & \texttt{arxiv 2022} & --- & --- & --- & \cmark \\ 
        DeepCut \cite{aflalo2022deepcut} & \ref{sec:training:head} & \cmark & \texttt{arxiv 2022} & \cmark & \cmark & --- & --- \\ 
        UMOD \cite{kara2023umod} & \ref{sec:3:single-to-multi:iterative}, \ref{sec:training:task-specific} & --- & \texttt{WACV 2023} & --- & --- & \cmark & \cmark \\ 
        FOUND \cite{simeoni2023found} & \ref{sec:found-coarse}, \ref{sec:training:head} & \cmark & \texttt{CVPR 2023} & \cmark & \cmark & --- & --- \\ 
        CutLER \cite{wang2023cutler} & \ref{sec:3:single-to-multi:iterative}, \ref{sec:training:task-specific} & \cmark & \texttt{CVPR 2023} & --- & --- & \cmark & \cmark \\ 
        Ex.-FreeSOLO \cite{ishtiak2023examplar-freesolo} & \ref{sec:training:task-specific} & --- & \texttt{CVPR 2023} & --- & --- & \cmark & \cmark \\ 
        WSCUOD \cite{lv2023wscuod} & \ref{sec:training:finetuning} & \cmark & \texttt{arxiv 2023} & \cmark & \cmark & \cmark & --- \\ 
        UCOS-DA \cite{zhang2023ucos-da} & \ref{sec:training:head} & \cmark & \texttt{ICCVW 2023}  &\cmark & --- & --- & --- \\ 
        MOST \cite{rambhatla2023most} & \ref{sec:3:single-to-multi:adaptive} & \cmark & \texttt{ICCV 2023} & \cmark & \cmark & \cmark & --- \\ 
        SEMPART \cite{ravindran2023sempart} & \ref{sec:training:head} & --- & \texttt{ICCV 2023} & \cmark & \cmark & --- & --- \\ 
        Box-based \cite{gomel2023boxbasedrefinement} & \ref{sec:training:finetuning} & \cmark & \texttt{ICCV 2023} & --- & \cmark & --- & --- \\ 
        UOLwRPS \cite{song2023UOLwRPS} & \ref{sec:training:head} & \cmark & \texttt{ICCV 2023} & --- & \cmark & --- & --- \\ 
        PaintSeg \cite{li2023paintseg} & \ref{sec:pixel-refinement} & --- & \texttt{NeurIPS 2023} & \cmark & --- & --- & --- \\ 
    \bottomrule
    \end{tabular}
    }
    \caption{\label{tab:method_overview}  
    \textbf{Overview of unsupervised object localization literature}. Presentation of the different methods 
    discussed in this survey paper. We specify the section where they are discussed (column `Sec'), if the code is available, the venue where they have been published (if applicable) and the specific tasks they have been implemented for and tested on.
    Superscript indicates which papers evaluated the method in the setting.} 
\end{table*}

\begin{table*}
    \centering
    \resizebox{\linewidth}{!}{
    \rowcolors{2}{gray!15}{white}
    \begin{tabular}{@{} p{0.28\linewidth} l p{0.25\linewidth} p{0.2\linewidth} p{0.35\linewidth} @{}}
    \toprule
        \textbf{Task} & \textbf{Sec.} & \textbf{Short description} & \textbf{Standard metrics} & \textbf{Classical evaluation datasets} \\ 
    \midrule
        \Unsupsaliency & \ref{sec:background:def:unsup-saliency} & Segment foreground & IoU, Acc, max F$_\beta$ & DUT-OMRON \cite{yang2013dut-omron}, DUTS-TE \cite{wang2017duts-te}, ECSSD \cite{shi2016ecssd}, CUB-200-2011 \cite{wah2011cub} \\ 
        \Singleobjectdiscovery & \ref{sec:background:def:single-object} & Detect one main object & CorLoc & PASCAL VOC07 \cite{pascal-voc-2007}, PASCAL VOC12 \cite{pascal-voc-2012}, COCO 20k \cite{coco2014,vo2021large_scale_unsup_object_discovery} \\ 
        \Multiclassif{} & \ref{sec:background:def:multi-object-classif} & Detect all objects & AP$_{50}$, AP$_{75}$, AP, odAP$_{50}$, odAP & COCO 20k \cite{coco2014,vo2021large_scale_unsup_object_discovery}, PASCAL VOC07 \cite{pascal-voc-2007}, PASCAL VOC12 \cite{pascal-voc-2012}, COCO val2017 \cite{coco2014}, UVO \cite{wang2021uvo} \\ 
        \Multiseg{} & \ref{sec:background:def:multi-object-segmentation} & Segment all objects & AP$_{50}$, AP$_{75}$, AP, mIoU & COCO 20k \cite{coco2014,vo2021large_scale_unsup_object_discovery}, COCO val2017 \cite{coco2014}, VOC12 \cite{pascal-voc-2012}, UVO \cite{wang2021uvo}\\ 
    \bottomrule
    \end{tabular}
    }
    \caption{\label{tab:tasks}\textbf{Evaluating unsupervised object localization:} tasks, metrics and typical evaluation datasets.}
\end{table*}

\section{Problem definition: tasks \& metrics}
\label{sec:background}

\PP{We present in this section, the background material for the tasks of interest in this paper.} 
When talking about \emph{unsupervised object localization}, four typical tasks are considered: First, \emph{\unsupsaliency{}} (1) aims at separating 
foreground objects from the background; \emph{\singleobjectdiscovery{}} (2) requires to localize one object per image with a box, while all objects must be detected when performing \emph{\multiclassif{}} (3); finally \emph{\multiseg{}} (4) demands instance masks for all objects. These tasks are illustrated in \autoref{fig:unsup-tasks} and their evaluation protocols, along with typical datasets and metrics, are detailed in \autoref{tab:tasks}.

\subsection{Notations}
If not otherwise stated, object localization is performed on a single RGB image $\mathbf{X} \in \mathbb{R}^{W \times H \times 3}$ of width $W \in \mathbb{N}$ and height $H \in \mathbb{N}$. 

\subsection{\Unsupsaliency{}}
\label{sec:background:def:unsup-saliency}

\textbf{\textit{Task Description}}. 
This task involves processing an input image $\mathbf{X}$ with the aim of generating a foreground/background binary mask $\mathbf{m} \in \{0, 1\}^{W \times H}$. This binary mask is computed in such a way that each pixel location is assigned a value of 1 if it belongs to the foreground object(s), and 0 otherwise, effectively highlighting the object(s) of interest in the image. Note that this task is sometimes referred as `single-object segmentation' \cite{aflalo2022deepcut}. See \autoref{fig:foreground-task} for an illustration.

\parag{Metrics}. 
Methods are evaluated with:
\begin{itemize}
    \item The intersection-over-union (\textbf{IoU}), which measures the overlap of foreground regions between the predicted and the ground-truth masks, averaged over the entire dataset;
    \item The pixel accuracy (\textbf{Acc}), which measures the pixel-wise accuracy between the predicted binary mask $\mathbf{m}$ and the ground-truth mask;
    \item The maximal $F_\beta$ score (\textbf{max\,$F_\beta$}),
    where $F_\beta$ is a weighted harmonic mean of the precision (P) and the recall (R) between the predicted mask $\mathbf{m}$ and the ground-truth mask:
    \begin{equation}
        F_\beta = \frac{(1 + \beta^2) \text{P} \times \text{R}}{ \beta^2\text{P} + \text{R}}.
    \end{equation}
    The value of $\beta$ is generally set to $\beta^2=0.3$ following \cite{wang2022tokencut,shin2022selfmask,simeoni2023found}.
    $F_\beta$ is computed over masks which have binarized with different thresholds between $0$ and $254$. The max\,$F_\beta$ metric finds the optimal threshold over the whole dataset that yields the highest $F_\beta$ for all the generated binary masks.
\end{itemize}

\noindent\textbf{\textit{Evaluation datasets}}. Unsupervised saliency detection is typically evaluated on a collection of datasets depicting a large variety of objects in different backgrounds. Popular saliency datasets are: DUT-OMRON \cite{yang2013dut-omron} (5,168 images), DUTS-TE \cite{wang2017duts-te} (5,019 test images), ECSSD \cite{shi2016ecssd} (1,000 images), and CUB-200-2011 \cite{wah2011cub} (1,000 test images).
Unsupervised methods that require a self-training step (e.g., \cite{wang2022freesolo,shin2022selfmask,bielski2022move}) are generally trained on DUTS-TR \cite{wang2017duts-te} (10,553 images).

\subsection{\Singleobjectdiscovery{}}
\label{sec:background:def:single-object}

\noindent\textbf{\textit{Task Description.}}
The primary objective of this task is to tightly enclose the main object or one of the main objects of interest 
within a bounding box.
For an illustration of this task see \autoref{fig:single-obj-task}.

\parag{Metrics.}
As in \cite{simeoni2021lost,wang2022tokencut,vo2021large_scale_unsup_object_discovery}, the Correct Localization (\textbf{CorLoc}) metric is reported. It measures the percentage of correct boxes, i.e., predicted boxes having an intersection-over-union greater than $0.5$ with one of the ground-truth boxes.

\parag{Evaluation datasets.}
Methods are typically evaluated on the \texttt{trainval} sets of PASCAL VOC07 \& VOC12 datasets which generally contain images with only a single large object \cite{pascal-voc-2007, pascal-voc-2012} and COCO20K (a subset of $19,817$ randomly chosen images from the COCO2014 trainval dataset \cite{coco2014} following \cite{vo2020unsup_multi_object_discovery, vo2021large_scale_unsup_object_discovery}) which includes images containing several objects.

\subsection{\Multiclassif{}}
\label{sec:background:def:multi-object-classif}

\noindent\textbf{\textit{Task Description.}} 
In contrast to \singleobjectdiscovery{}, 
the \multiclassif{} task
aims to detect and localize each individual object present in the image, regardless of class considerations.
Given an input image $\mathbf{X}$, the aim of the multi-object discovery task is to generate a set of bounding boxes. 
Note that this task is sometimes referred as `multi-object discovery' or `zero-shot unsupervised object detection' \cite{wang2023cutler}.
For an illustration of this task see \autoref{fig:multi-obj-task}.

\parag{Metrics.}
Methods are typically evaluated using the standard Average Precision (\textbf{AP}) metric that assesses detection precision across various confidence thresholds, quantified by the area under the precision-recall curve. Precision is the ratio of correct detection to all detection, while recall denotes the ratio of correct detection to all ground-truth objects in the dataset.

\OS{All predictions are sorted given an `objectness' score and are iteratively defined as `correct'} when the Intersection-over-Union (IoU) with \OS{an unassigned} ground-truth object exceeds a specified threshold.
\OS{In this setup,} the specific class of the ground-truth object (if available) does not affect the matching process.
Typical IoU thresholds for correct detection are 0.5 (\textbf{AP$_{50}$}) or 0.75 (\textbf{AP$_{75}$}). AP can be computed at various IoU thresholds from 0.5 to 0.95 with 0.05 intervals, and then averaged (sometimes denoted AP@[50-95] or more simply \textbf{AP}).
AP can also be calculated separately for small (\textbf{AP$_S$}), medium (\textbf{AP$_M$}) and large (\textbf{AP$_L$}) sized objects specifically.
Additionally, there is the less common \textbf{odAP} metric, introduced by \cite{vo2021large_scale_unsup_object_discovery}, which averages AP values for detected objects at each number of detection, from one to the maximum number of ground-truth objects in an image within the dataset; odAP is thus independent of the number of detection per image.
Some works \cite{wang2022freesolo,ishtiak2023examplar-freesolo} also report the Average Recall (\textbf{AR$_k$}) which measures the maximum recall for a fixed number $k$ of detection per image, being more permissive to redundant and random detection results than AP.

\parag{Evaluation datasets.}
Models are typically evaluated on the \texttt{trainval} sets of PASCAL VOC07 \cite{pascal-voc-2007} \& VOC12 \cite{pascal-voc-2012} datasets and COCO 20k \cite{coco2014,vo2021large_scale_unsup_object_discovery}, following \cite{simeoni2021lost}, or on the \texttt{test} set of PASCAL VOC07 \cite{pascal-voc-2007} and the \texttt{validation} set COCO \cite{coco2014}, following \cite{wang2023cutler}.
Some works also evaluate on the \texttt{val} split of the Unidentified Video Object (UVO) dataset \cite{wang2021uvo}.

\subsection{\Multiseg{}}
\label{sec:background:def:multi-object-segmentation}

\noindent\textbf{\textit{Task Description.}}
\Multiseg{} involves analyzing an input image $\mathbf{X}$ to generate a set of binary masks $\mathbf{M} = \{\mathbf{m}_i\}_{i=1}^{N_B}$, where each $\mathbf{m}_i \in \{0, 1\}^{W \times H}$ represents a binary foreground/background segmentation mask for the $i$-th detected object, with $N_B$ being the total number of detected objects.
Unlike \multiclassif{} which 
localizes objects through bounding boxes, this task 
generates pixel-level masks for each individual object. %
Note that 
it is sometimes referred as `zero-shot unsupervised instance segmentation' \cite{wang2023cutler}.
For an illustration of this task see \autoref{fig:instance-seg}.

\parag{Metrics.}
Similarly to the \multiclassif{} task, Average Precision is reported at various IoU thresholds (\textbf{AP$_{50}$}, \textbf{AP$_{75}$}, and \textbf{AP}) between predicted and ground-truth masks.
Additionally, some works \cite{kim2020unsupervised,kara2023umod} calculate the mIoU as the average IoU between each ground-truth mask (including the background) and the detected mask with the highest IoU.

\parag{Evaluation datasets.}
The task is typically evaluated on the \texttt{trainval} or the \texttt{validation} set of COCO \cite{coco2014}, following \cite{wang2023cutler}, but also on COCO 20k \cite{coco2014,vo2021large_scale_unsup_object_discovery}, Pascal VOC12 \cite{pascal-voc-2012}, and less commonly on the \texttt{val} set of the UVO dataset \cite{wang2021uvo}.

\section{Training-free object localization with self-supervised ViTs}
\label{sec:ssonly}

\begin{figure*}[ht!]
    \setlength{\tabcolsep}{0.2pt}
    \renewcommand{\arraystretch}{0.2}
    
    \centering
     \begin{subfigure}[b]{0.6\textwidth}
        \centering
        \begin{tabular}{ccc}
           \includegraphics[width=0.3\textwidth]{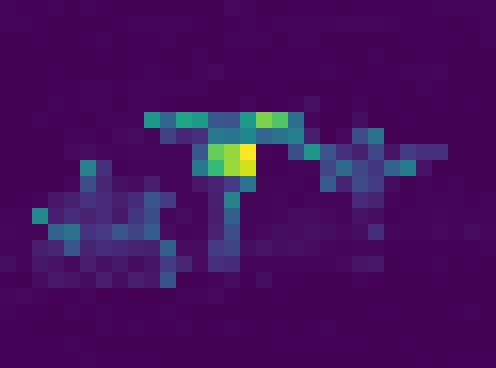}   & 
           \includegraphics[width=0.3\textwidth]{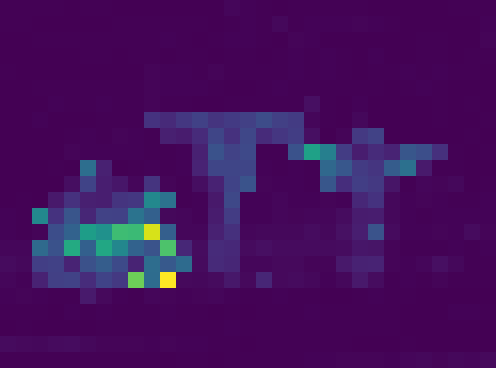} & 
           \includegraphics[width=0.3\textwidth]{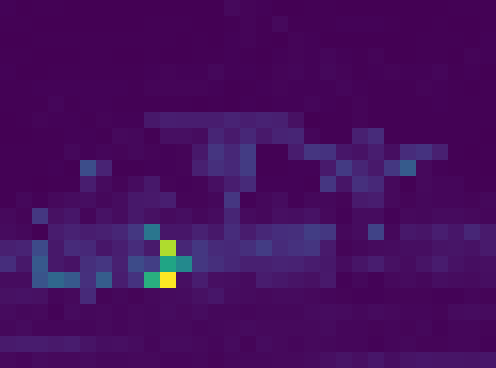}  
           \\
           \includegraphics[width=0.3\textwidth]{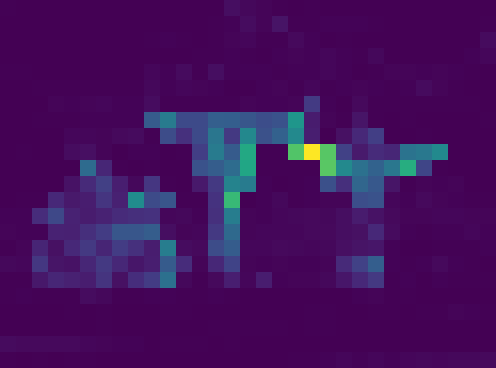} 
           &
           \includegraphics[width=0.3\textwidth]{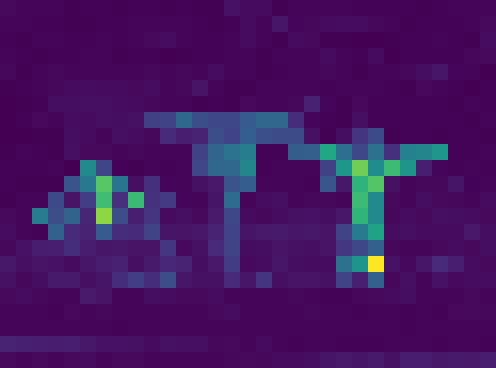}   & 
           \includegraphics[width=0.3\textwidth]{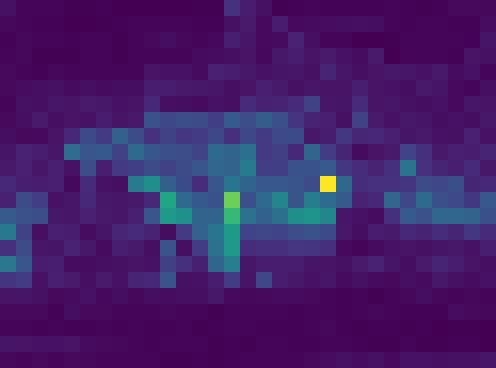}
        \end{tabular}
        \caption{\centering \cls attention (six heads) \textcolor{white}{AAAAAAAAAAAAAAAAAAAAAAAAAA} 
        }
        \label{fig:feats-cls}
     \end{subfigure}%
     \hfill
     \begin{subfigure}[b]{0.4\textwidth} 
        \centering
        \begin{tabular}{cc}
           \includegraphics[width=0.5\textwidth,height=2.12cm]{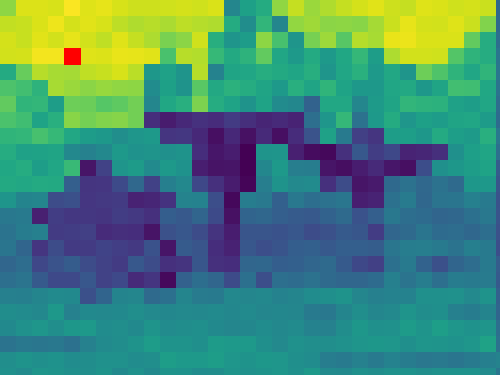}   &
           \includegraphics[width=0.5\textwidth,height=2.12cm]{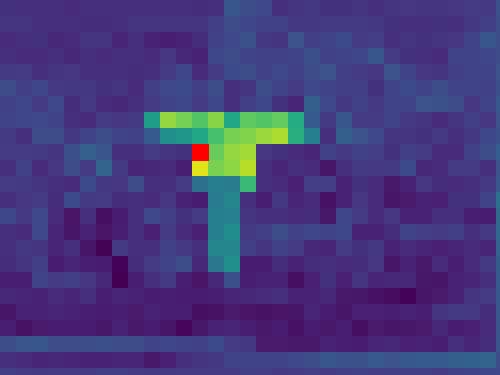} \\
           \includegraphics[width=0.5\textwidth,height=2.12cm]{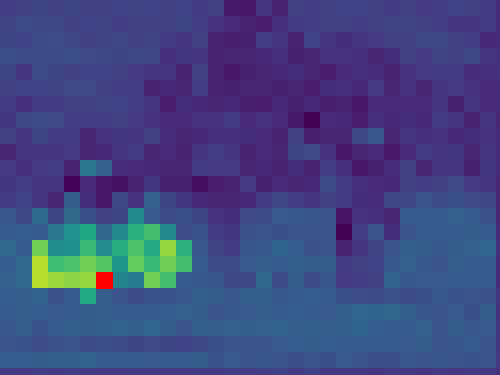}   & 
           \includegraphics[width=0.5\textwidth,height=2.12cm]{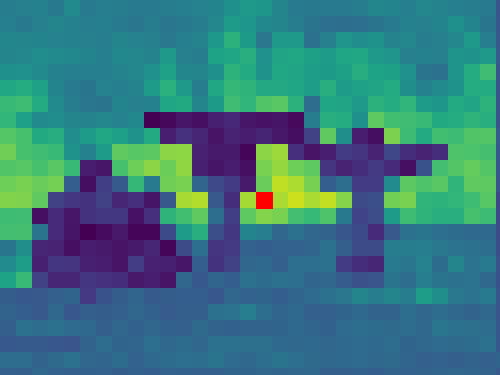} 
        \end{tabular}
        \caption{\centering Patch correlation to a patch (\textcolor{red}{red} square) given the \emph{key} feats.}
        \label{fig:feats-corr}
     \end{subfigure}

    \centering
     \begin{subfigure}[b]{0.32\textwidth}
        \centering
        \includegraphics[width=0.8\textwidth]{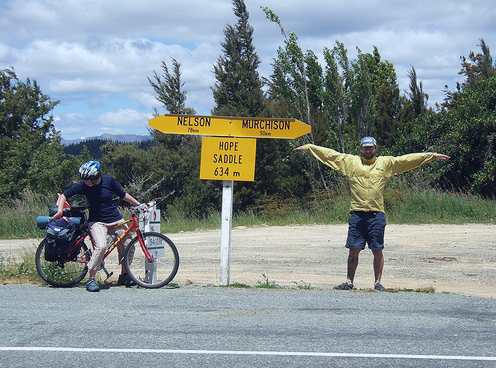}
        \caption{\centering Original image 
        \textcolor{white}{AAAAAAAAAAAAA}}
    \end{subfigure}%
    \hfill
     \begin{subfigure}[b]{0.32\textwidth}
        \centering
        \includegraphics[width=0.8\textwidth]{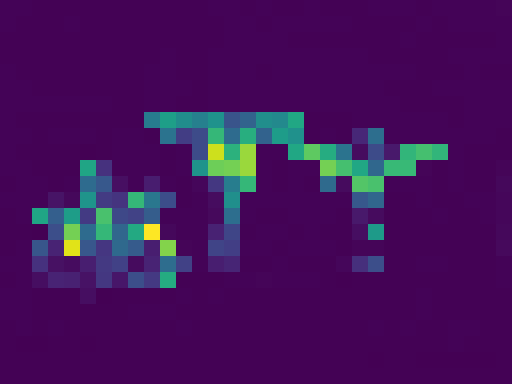}%
        \caption{\centering Inverse degree 
            (used\,in\,LOST~\cite{simeoni2021lost})}
    \end{subfigure}%
    \hfill
     \begin{subfigure}[b]{0.32\textwidth}
          \centering
          \includegraphics[width=0.8\textwidth]{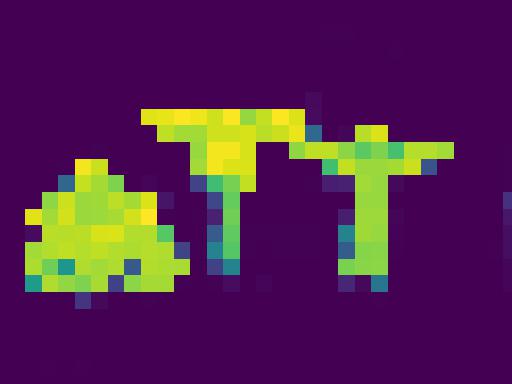}%
          \caption{\centering 2nd eigenvector (used\,in\,TokenCut\,\cite{wang2022tokencut})}
        \label{fig:feats-corr2}
     \end{subfigure}
     
    \caption{\textbf{Object localization using DINO's last attention layer}.
    Visualization of different features extracted from the last attention layer of DINO~\cite{caron2021dino} for the original image (c):
    (a) \cls attention maps generated with all heads; 
    (b) Correlation between a patch of interest (in \textcolor{red}{red}) and all other patches given the \emph{key} features of the last MSA layer; 
    (d-e) Inverse degree matrix and second eigenvector, which are used to extract finer localization information \cite{simeoni2021lost, wang2022tokencut} (more details in \autoref{sec:no-training:degree}, \ref{sec:no-training:clustering}).
    }
    \label{fig:dino}
\end{figure*}

Recent advances on visual transformers and self-supervised learning have paved the way to efficient unsupervised object localization strategies discussed in this section. We first provide notations in \autoref{sec:feats:notations}, introduce the opportunity offered by self-supervised features in \autoref{sec:feats:self}. \OS{In \autoref{sec:no-training} we discuss ways to exploit those features to localize a single object per image, and several objects in \autoref{sec:3:single-to-multi}. We discuss different post-processing strategies in \autoref{sec:post-processing} that help improve the quality of localization. Finally, we quantitatively compare the different methods in \autoref{sec:quantitative-onlyfeats}. }

\subsection{Notations}
\label{sec:feats:notations}

Originally designed for the domain of natural language processing, \emph{transformers} \cite{vaswani2017attentionisallyouneed} have been effectively adapted to the domain of computer vision \cite{dosovitskiy2021vit}.
\OS{Vision Transformers (ViTs) consider an input image $\vX \in \PP{\mathbb{R}^{W \times H \times 3}}$ as a sequence of $N$ tokens, each token corresponding to a local image patch of a fixed size $P \times P$.}
For each patch, ViT models generate a patch embedding of dimension $d$ using a trainable linear projection layer followed by a position embedding to preserve positional information.
A learnable embedding called the `class token', noted \cls, is appended to the patch embeddings, resulting in a transformer input of dimensionality $\mathbb{R}^{(N+1) \times d}$.

\OS{
Transformers process the input through a series of layers comprising multi-head self-attention (MSA) and multi-layer perceptron (MLP) blocks.}
\OS{While the role of the MLP blocks is to independently process each patch embedding, MSA blocks allow tokens to gather information from other tokens, enhancing contextual understanding.}
Specifically, in self-attention \cite{vaswani2017attentionisallyouneed}, the token embeddings are initially projected into three learned spaces, resulting in query ($\vQ$), key ($\vK$), and value ($\vV$) features, all residing in $\real^{(N + 1)\times d}$. 
Subsequently, the self-attention output is computed as $\vY =  \text{softmax}\left( d^{-1/2} \, \vQ \vK\tran \right) \vV \in \real^{(N+1) \times d}$, with the softmax operation applied row-wise.
\OS{Description here corresponds to} a single-head attention layer, whereas MSA blocks consist of multiple parallel self-attention heads, with their outputs concatenated and subsequently processed by a linear projection layer.

\subsection{Self-supervised features}
\label{sec:feats:self}

\OS{Alongside the developments of ViTs, self-supervised training strategies \cite{caron2021dino,he2022mae,zhou2022ibot,assran2022masked,chen2021mocov3} have successfully been designed to learn useful representations without any manual annotation.}
Interestingly, Caron et al.~\cite{caron2021dino} have shown that ViTs pre-trained in a self-supervised manner exhibit \emph{strong localization properties} which contrast with those of models trained using fully supervised methods for image classification. 
\OS{We visualize in \autoref{fig:feats-cls} the attention maps of the \cls token of each head and observe that foreground objects indeed receive most of the attention. }

Although the localization properties of the attention of DINO \cite{caron2021dino} are visually enticing, they hardly suffice to directly localize objects. Indeed, different heads of the ViT model focus on different objects and regions of the image. 
Moreover, complex scenes have noisier attention \cite{simeoni2021lost}. Therefore, it is not obvious \emph{what is object or not.}

\OS{A possible strategy to extract object localization information from the \cls attention maps consists in choosing the map of the \emph{best}} head, binarize it and get the connected components. The choice of the attention head can be fixed based on dataset performance or determined dynamically per image using heuristics, e.g., selecting the head producing the mask with the highest average IoU overlap with other heads' outputs. We refer to this strategy as `DINO' in tables. Although this approach is straight-forward, it does not fully exploit the potential of the pre-trained ViT as shown in \cite{simeoni2021lost}. 
Amir et al.~\cite{amir2021deep} also reveal that self-supervised features have a rich semantic space which allows them to easily find object parts shared amongst different semantically close objects, for instance animals.

\OS{Moving the focus from the attention maps to the features of the last MSA layer, different works \cite{simeoni2021lost,wang2022tokencut,wang2023cutler} have shown that in particular the \emph{key} features have very good correlation properties (as visible in \autoref{fig:feats-corr}) and propose simple strategies to extract objects, which we describe now.}

\subsection{Training-free single-object localization with ViT self-supervised features}
\label{sec:no-training}

In this section, we discuss some of the latest unsupervised methods for locating a single object in an image. \gilles{One way to do so is to apply directly $k$-means on self-supervised features \cite{lim2022imst}, or to project such features on their first component after PCA analysis and use a simple threshold to separate objects and background \cite{lv2023wscuod}. We rather concentrate here on methods which leverage correlations among patches and are able to achieve high performance}
\emph{without needing any additional training step}.
\OS{Most of those methods}
are based on the following key observations about self-supervised features (especially those 
of DINO \cite{caron2021dino}): (1) features from two different patches \PP{of a same object} 
are highly correlated; (2) features from two different background patches are highly correlated; but (3) features from an object patch and a background patch do not correlate well.
As a result, when constructing a similarity graph $\mathcal{G}$ among all patches in a image, where similarity is defined as feature correlation, object and background patches naturally form distinct clusters within this graph. This explains why recent methods leverage such a graph and different ways to identify an `object cluster' in $\mathcal{G}$.

\begin{figure}[ht!]
    \centering
    \includegraphics[width=\columnwidth]{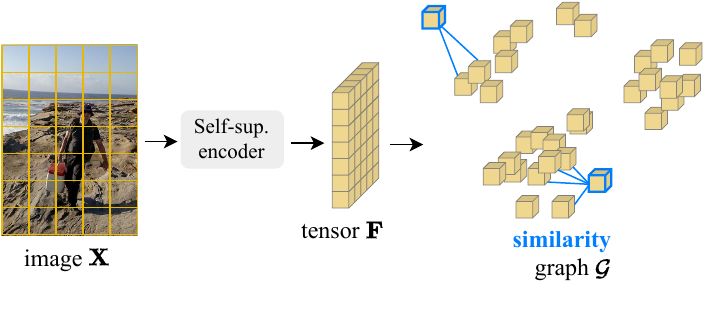}
    \caption{\textbf{Feature similarity graph for unsupervised object localization}.
    A similarity graph $\mathcal{G}$ among patches of an image is built and used by unsupervised object localization methods \cite{simeoni2021lost, wang2022tokencut, shin2022selfmask, wang2023cutler, simeoni2023found}.}
    \label{fig:graph}
\end{figure}

\begin{figure*}
     \centering
     \begin{subfigure}[b]{0.3\textwidth}
         \centering
         \includegraphics[width=.95\textwidth]{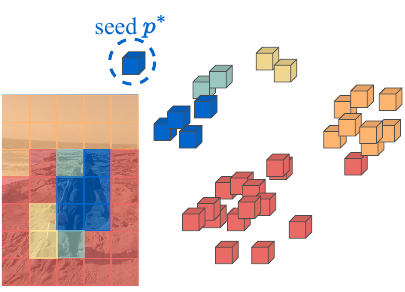}
         \caption{LOST \cite{simeoni2021lost}}
         \label{fig:lost}
     \end{subfigure}
     \hfill
     \begin{subfigure}[b]{0.3\textwidth}
         \centering
         \includegraphics[width=.95\textwidth]{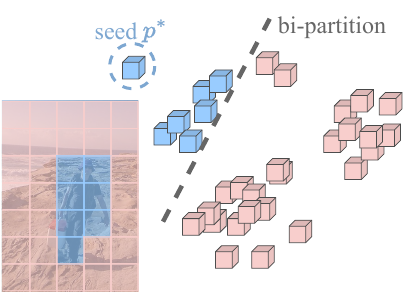}
         \caption{TokenCut \cite{wang2022tokencut}}
         \label{fig:tokencut}
     \end{subfigure}
     \hfill
     \begin{subfigure}[b]{0.3\textwidth}
         \centering
         \includegraphics[width=\textwidth]{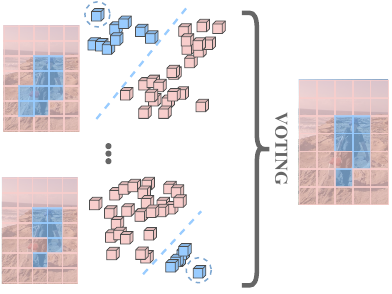}
         \caption{SelfMask \cite{shin2022selfmask} coarse mask}
         \label{fig:selfmask}
     \end{subfigure}
        \caption{\textbf{Different strategies to exploit the similarity graph $\mathcal{G}$.} (a) LOST \cite{simeoni2021lost} exploits the inverse degree information to find the seed patch $p^*$ (from high in \textcolor{RoyalBlue}{blue} to low in \textcolor{BrickRed}{red}); (b) TokenCut \cite{wang2022tokencut} splits the graph in two subsets given a bipartition (in \textcolor{CornflowerBlue}{blue} and \textcolor{Lavender}{pink}); (c) SelfMask~\cite{shin2022selfmask} computes the partitioning on different features before voting for the most popular. Those methods are discussed in details in \autoref{sec:no-training:degree}, \autoref{sec:no-training:clustering} and \autoref{sec:3:all-multi}.}
        \label{fig:three graphs}
\end{figure*}

\subsubsection{Patch-similarity graph}
\gilles{The graph $\mathcal{G}$ introduced before is constructed using 
$\ell_2$-normalized patch features} ${\vf_{p} \in \real^{d}}$
with $p \in \{1, \ldots, N\}$ \OS{the index corresponding to each patch position ($N$ is the total number of patches).}
\OS{Most} methods below leverage ViTs and this patch feature is typically the normalized \emph{key} of patch $p$ in the last attention layer--\OS{which has been shown to have the best correlation properties \cite{simeoni2021lost,wang2022tokencut}}. The undirected graph of patch similarities $\mathcal{G}$ is then represented by the binary (symmetric) adjacency matrix $\vA \,{=}\, (a_{pq})_{1 \leq p, q\leq N} \in \{0, 1\}^{N \times N}$ such that
\begin{align}
\label{eq:sim}
a_{pq} = 
\left\{
    \begin{array}{ll}
    1 & \text{if } \vf_p^{\smash{\tran}} \vf_q \geq \delta, \\
    \varepsilon & \text{otherwise},
    \end{array}
\right.
\end{align}
with $\delta \in [0, 1]$ a constant threshold and $\varepsilon \geq 0$ a small value ensuring a full connectivity of the graph.
In this graph, two patches $p$ and $q$ are thus connected by an undirected edge if their features $\vf_p$ and $\vf_q$ are sufficiently positively correlated.
We review below the different ways to extract an object cluster from the graph encoded by $\vA$.

\subsubsection{Growing object clusters from selected patch seeds}
\label{sec:no-training:degree}
Beyond the correlation properties listed above, \OS{it is possible to highlight a single cluster in $\mathcal{G}$ by exploiting the following assumption:} 
objects often occupy less space than the background \OS{in images}, hence object patches tend to have a smaller number of connection in $\mathcal{G}$ than background patches. \OS{Leveraging both assumptions}, LOST \cite{simeoni2021lost} \OS{measures} this quantity 
by using the \emph{degree} of the nodes in $\mathcal{G}$: 
\begin{equation}
\label{eq:degree}
d_p = \sum_{q=1}^{N} a_{pq}.
\end{equation}
The patch ${p^* = \arg \min d_p}$ with the smallest degree in $\mathcal{G}$ is \OS{selected as likely corresponding to}
an object patch and is called the \emph{seed}. We visualize the seed selection in \autoref{fig:lost}. Then all patches connected to $p^*$ in $\mathcal{G}$ are considered part of the same object. Note that $\delta = \varepsilon = 0$ in LOST.

Because the process above tends to miss some parts of the object, an extra step of \emph{`seed expansion'} is proposed in LOST. It
consists in selecting the next best seeds after $p^*$. 
\OS{They are defined} as the patches of small degree that are connected to $p^*$ in $\mathcal{G}$: ${\mathcal{S} = \{s \mid s \in \mathcal{D}_k \text{ and } \vf_s^{\tran} \vf_{p^*} \geq 0\}}$ where $\mathcal{D}_k$ is the set of the $k$ patches with the smallest degrees, with $k$ a hyper-parameter with a typical value $k=100$. The final \OS{binary} object mask $\vm = (m_q)_{1 \leq q \leq N}$ \OS{highlights}
all patches that, on average, correlate positively with one of the seed patches:
\begin{align}
\label{eq:lost}
m_q = 
\left\{
    \begin{array}{ll} 
    1 & \text{if } \sum_{s \in \mathcal{S} } \vf_q^{\tran} \vf_{s} \geq 0, \\
    0 & \text{otherwise}.
    \end{array}
\right.
\end{align}

Subsequent MaskDistill~\cite{gansbeke2022maskdistill} employs a similar process to
identify an object cluster 
but starts
from patch seeds that are selected 
using the attention maps with the \cls token at the last layer of a ViT, instead of relying on the patch degree information. 

\subsubsection{Finer localization with spectral clustering}
\label{sec:no-training:clustering}

Instead of identifying one seed in an object cluster and a cluster in which this seed belongs, TokenCut \cite{wang2022tokencut} \OS{and DeepSpectralMethods \cite{melas2022deepsectralmethod} use} 
spectral clustering to separate objects and background. The first considers an
adjacency matrix $\vA$ computed using $\delta=0.2$ and $\varepsilon=10^{-5}$, while the second uses $\delta=\varepsilon=0$ and combines it with a nearest neighbors adjacency matrix based on color affinity. Both compute the second smallest eigenvalue and the corresponding eigenvector $\indicatory^*$ to the generalized eigensystem 
\begin{align}
\label{eq:eigenproblem}
(\vD - \edge) \indicatory = \lambda \, \vD \indicatory,
\end{align}
where $\vD$ is the diagonal degree matrix with entries $d_p$ (see Eq.\,\ref{eq:degree}). The patches are then partitioned into two clusters $\mathcal{A} = \{p \, | \, \indicatory^*_p \le \overline{\indicatory}^*  \}$ and $\mathcal{B} = \{p \, | \, \indicatory^*_p > \overline{\indicatory}^* \}$, where $\overline{\indicatory}^*$ is the average of the entries in $\indicatory^*$ \gilles{in \cite{wang2022tokencut} and $\overline{\indicatory}^* = 0$ in \cite{melas2022deepsectralmethod}}. We visualize the bi-partition in \autoref{fig:tokencut}. The connected component (in the spatial domain)
with the main salient object is identified as the one containing the largest absolute value $\max_p \abs{\indicatory^*_p}$ in \cite{wang2022tokencut}, while \cite{melas2022deepsectralmethod} select the smallest component.  
All patches in this component define the object mask $\vm$.

\subsection{From single- to multi-object localization}
\label{sec:3:single-to-multi}
Methods described so far are limited to the discovery of one object. Recent methods, like \cite{wang2022freesolo,simeoni2023found,wang2023cutler}, address this limitation and are able to discover multiple objects in one image. 
We discuss first how to discover several objects using clustering-based strategies inspired by TokenCut \cite{wang2022tokencut} (\autoref{sec:3:single-to-multi:iterative}) or with 
solutions which aim at discovering the different objects \emph{seeds} at once (\autoref{sec:3:single-to-multi:adaptive}).

\subsubsection{Iterative clustering-based multi-object discovery}
\label{sec:3:single-to-multi:iterative}

\begin{figure}[ht!]
    \centering
    \includegraphics[width=0.98\columnwidth]{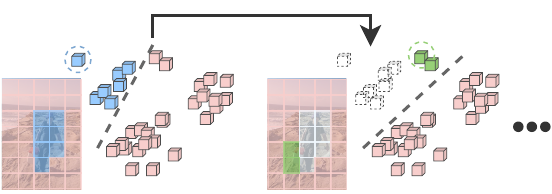}
    \caption{\OS{\textbf{Iterative selection in MaskCut \cite{wang2023cutler}}}. Object patches are iteratively selected in $\mathcal{G}$.}
    \label{fig:maskcut}
\end{figure}
In an attempt to adapt TokenCut~\cite{wang2022tokencut} to multi-objects localization, Wang et al. \cite{wang2023cutler} propose to compute the first mask $\vm$ and disconnect the corresponding patches in the graph $\mathcal{G}$ before re-applying spectral clustering to discover a second object. This simple process \OS{(visualized in \autoref{fig:maskcut})} is repeated a pre-determined number of times (3 in practice), \OS{named MaskCut}, allowing the discovery of a few well localized objects.
\OS{Also}, rather than strictly disconnecting the patches 
\OS{of} $\vm$ in $\mathcal{G}$, the authors set a small edges weight $a_{pq} = \varepsilon$ if $\vm_p = 1$ or $\vm_q = 1$ and then compute the second eigenvector of the system \eqref{eq:eigenproblem} one more time.

Similarly, UMOD \cite{kara2023umod} also builds upon TokenCut. %
It computes the eigenvectors associated to the $k$ smallest eigenvalues of the system \eqref{eq:eigenproblem}. These eigenvectors can be gathered in a matrix $\vY \in \mathbb{R}^{N \times k}$ and the masks are obtained by applying $k$-means on the rows of $\vY$, as in SelfMask \cite{shin2022selfmask}. The background cluster is identified as the one covering the largest area in the image, as assumed in LOST \cite{simeoni2021lost}, while the remaining clusters 
represent `object areas'.
Distinct from CutLER, UMOD \cite{kara2023umod} dynamically determines the optimal number of clusters using an iterative process.
Practically, the number of clusters is incremented from 2 until the relative change in total object area gets below a predefined threshold.

\subsubsection{Discovering multiple objects at once}
\label{sec:3:single-to-multi:adaptive}

\parag{Entropy-based multi-object discovery.}
Alternatively, 
MOST~\cite{rambhatla2023most} exploits the correlation maps between the patch features. The authors notice that \OS{such} maps are sparser when computed using an object patch as reference than when using a background patch. Thus they propose to analyze all these maps to determine a set of patches likely to fall on objects. The correlation maps are processed with an Entropy-based Box Analysis (EBA) method which discriminates the correlation maps based on their entropy (measured at multiple scales). The patches that yield the correlation maps with the lowest entropy (likely to be object patches) are clustered providing different \emph{pools}. One object mask is then created from each of these pools with a process similar to LOST~\cite{simeoni2021lost}: (1) the patches of the lowest degree in $\mathcal{G}$ in the current pool are extracted, playing the role of the seed (called \emph{core} in \cite{rambhatla2023most}); (2) all patches in the current pool negatively correlated with the seed are filtered out; (3) the mask is then made of all patches positively correlated with the patches left in the pool. This three-step process is repeated for each pool, hence discovering multiple objects.

\begin{figure}[h!]
    \centering
    \begin{tabular}{cc}
         \includegraphics[height=4cm]{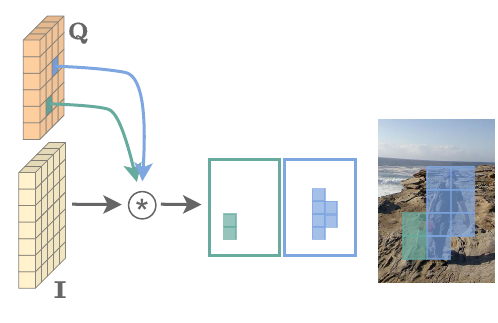} 
    \end{tabular}
    \caption{\textbf{Generation of several potential masks in FreeMask~\cite{wang2022freesolo}}. As many masks are produced as there are patches in $\vQ$, tensors \OS{$\vQ$ and $\vI$ being features extracted from a ResNet50 pre-trained with self-supervised denseCL \cite{wang2021densecl}.}}
    \label{fig:freemask}
\end{figure}

\parag{Query-based multi-object discovery.}
Leveraging pixel-level similarity, FreeMask~\cite{wang2022freesolo} exploits this time convolutional neural networks -- a ResNet50 \cite{he2016residual} pre-trained with self-supervised denseCL \cite{wang2021densecl}; \PP{Its principle is illustrated in \autoref{fig:freemask}}. 
\PP{The features $\vI \in \real ^{H \times W \times d}$, generated for the image $\vX$, are down-sampled into features $\vQ \in \real ^{H' \times W' \times d}$ with reduced sizes $H'$ and $W'$.} 
\gilles{The downsampled features $\vQ$ are reshape to matrix of size $N' \times d$ where $N' = H'W'$. We denote by $\vQ_q \in \mathbb{R}^{d}$ the features at spatial position $q$ in $\vQ$.}
\gilles{Each $\vQ_q$} is then considered as a seed and compared to all pixels in $\vI$ producing a soft map 
$\Vs \in \real^{H \times W \times N'}$
with entries
\begin{equation}
    s_{i,j,q} = \frac{\vQ_q^{\tran} \vI_{i,j}}{\|\vQ_q\| \xspace \|\vI_{i,j}\|}, 
\end{equation}
which is normalized to range in $[0,1]$. %
\gilles{This soft map is then binarized to obtain $\vm  \in \real^{H \times W \times N'}$ with entries}
\begin{align}
\label{eq:mask_graph_cut}
m_{i,j,q} = 
\left\{
    \begin{array}{ll} 
    1 & \text{if } s_{i,j,q} > \tau, \\
    0 & \text{otherwise},
    \end{array}
\right.
\end{align}
with $\tau\!\in\, (0,1]$ a hyper-parameter. 
\gilles{These $N'$ masks of spatial size $H \times W$ in $\vm$ are}
then sorted based on an \emph{objectness} score and ``the best'' ones are selected using an NMS-like function. In addition, different scales are used to produce more queries.

\subsection{Foreground/background separation}
\label{sec:3:all-multi}

As discussed above, discovering multiple objects using self-supervised features brings the challenge to assess the number of objects present in an image (either fixed as a hypothesis \cite{wang2023cutler} or discovered with a hand-crafted strategy \cite{rambhatla2023most, kara2023umod}). However, the task of foreground/background segmentation does not require to separate objects and can therefore be performed with simpler strategies.

\parag{Using spectral clustering for foreground segmentation.}
We have seen how spectral clustering on self-supervised features can be exploited to discover objects in \autoref{sec:no-training:clustering} and \autoref{sec:3:single-to-multi:iterative}. The same technique can be exploited for foreground segmentation, as done in SelfMask~\cite{shin2022selfmask}.
In this work, the authors rely on a pool of self-supervised backbones to extract an ensemble of plausible foreground/background masks. Concretely, SelfMask~\cite{shin2022selfmask} produces $k$ masks with \emph{each backbone} by: (a) computing the eigenvectors associated to the $k$ smallest eigenvalues in \eqref{eq:eigenproblem} (this time $a_{pq} = \vf_p^\intercal \vf_q$) and gather them in $\vY \in \mathbb{R}^{N \times k}$; and (b) producing the $k$ mask candidates by applying $k$-means on $\vY$. Given the mask candidates generated with the different pre-trained backbones, the \emph{most popular} mask $\vm$ is selected as the one with the highest average pairwise IoU with respect to all the other mask candidates. It is to be noted that before choosing $\vm$ the authors try and filter out likely wrong masks, e.g., with height (resp. width) as large as the original image height (resp. width). We visualize SelfMask process in \autoref{fig:selfmask}.

\begin{figure}[ht!]
    \centering
    \begin{tabular}{cc}
         \includegraphics[height=4cm]{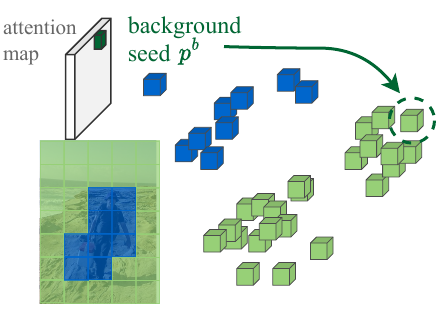} 
    \end{tabular}
    \caption{\OS{\textbf{Background discovery in FOUND \cite{simeoni2023found}.} Patches that are highly correlated to the background seed patch $p^b$ (the one receiving the least attention in the \cls attention maps) are identified as background. 
    See \autoref{sec:found-coarse}} for details.}
    \vspace{-10pt}
    \label{fig:found}
\end{figure}

\parag{\ttfamily{objects := not(background).}}
\label{sec:found-coarse}
Instead of constructing a method that specifically searches for a subset of the objects in the image, 
the authors of FOUND~\cite{simeoni2023found} propose to look at the problem the other way around and identify all background patches, hence discovering all object patches as a by-product with no need for a prior knowledge of the number of objects or their relative size with respect to the background. The \OS{strategy}, named `FOUND-coarse', starts  
by identifying the patch which receives the least attention with the \cls token in the last layer of a DINO-pretrained ViT. This patch is called the background seed and its index denoted by $b$.
Then the background mask $\vm^{\rm bck}$ (visualized in \autoref{fig:found}) has entries $m_p^{\rm bck}$, $p=1, \ldots, N$, satisfying
\begin{align}
\label{eq:found_mask}
m_p^{\rm bck} = 
\left\{
    \begin{array}{ll} 
    1 & \text{if } \vf_p^{\tran} \vf_{b} \geq \tau, \\
    0 & \text{otherwise}.
    \end{array}
\right.
\end{align}
In practice, the patch features $\vf_p$ are made of the $\ell_2$-normalized keys at the last attention layer and $\tau=0.3$. Note that, for conciseness, we omitted the fact that an adaptive re-weighting technique is applied to the entries of $\vf_p$ and $\vf_{b}$ in \eqref{eq:found_mask}. This re-weighting takes into account the fact that the background appears more or less clearly depending on the attention heads.
Finally, the object mask $\vm$ is the complement of $\vm^{\rm bck}$.

\subsection{Post-processing}
\label{sec:post-processing}

To adapt \OS{the produced masks} to the final \OS{downstream} task, it is possible to leverage different post-processing methods. We first discuss in  \autoref{sec:pixel-refinement} ways to improve the quality of generated masks by using pixel-level information. Second, in the case of object discovery/detection, we briefly discuss \OS{in \autoref{sec:mask-to-box}} how to generate boxes given the generated masks.

\subsubsection{Pixel-level refinement}
\label{sec:pixel-refinement}
In order to further improve the quality of the produced masks, popular refinement strategies \cite{bilateralsolverbarron2016,krahenbuhl2011crf} have been successfully applied to fit masks to pixel-level information. Indeed Bilateral Solver (BS) \cite{bilateralsolverbarron2016} and Conditional Random Field (CRF) \cite{krahenbuhl2011crf} use the raw pixel color and positional information in order to refine the generated coarse masks. Requiring no specific training, it is to be noted that the application of such methods can be expensive and sometime hurts the quality of the mask as discussed in \cite{bielski2022move, simeoni2023found}.

Alternatively, recent PaintSeg~\cite{li2023paintseg} introduces a training-free model to estimate \OS{precise} object foreground-background masks from either a bounding box or a predicted coarse mask. The method iteratively refines masks by in-painting the foreground and out-painting the background, updating masks through image comparisons. In-painting uses a pre-trained diffusion generative model \cite{rombach2022high}, while for image comparison employs a pre-trained DINO model~\cite{caron2021dino}. \OS{Such method generates high quality masks and obtains very good results as discussed in \autoref{sec:quantitative-onlyfeats}.}

\subsubsection{From mask to box}
\label{sec:mask-to-box}

In order to produce bounding boxes given a localization mask, e.g., for the task of unsupervised detection, methods separate the mask in connected components and enclose each of them with the tightest box. In the case of unsupervised object discovery, they produce a tight box only for the biggest component \cite{simeoni2023found, shin2022selfmask} or the one including the object seed $p^*$ \cite{simeoni2021lost, wang2022tokencut}. 

\begin{table*}[ht!]
    \centering
    \resizebox{\linewidth}{!}{

\setlength{\tabcolsep}{1.8pt}
\rowcolors{2}{gray!15}{white}
\begin{tabular}{@{} l *{13}{c} @{}}
    \toprule
    & & \multicolumn{3}{c}{DUT-OMRON} & \multicolumn{3}{c}{DUTS-TE} & \multicolumn{3}{c}{ECSSD} & \multicolumn{3}{c}{CUB} \\ \cmidrule(lr){3-5}\cmidrule(lr){6-8}\cmidrule(lr){9-11} \cmidrule(lr){12-14}
    Method & post-proc. & Acc & IoU & max\,$F_\beta$ & Acc & IoU & max\,$F_\beta$ & Acc & IoU & max\,$F_\beta$ & Acc & IoU & max\,$F_\beta$ \\ 
    \midrule
    \multicolumn{14}{c}{\tabletitle{SoTA before self-supervised ViTs era}} \\
     E-BigBiGAN \cite{voynov2021biggan}  & & 86.0 & 46.4 & 56.3 & 88.2 & 51.1 & 62.4 & 90.6 & 68.4 & 79.7 & --- & --- & ---\\
     Melas-Kyriazi et al. \cite{melas2021}  & & 88.3 & 50.9 & --- & 89.3 & 52.8 & --- & 91.5 & 71.3 & --- & --- & --- & ---\\
    \midrule
    \multicolumn{14}{c}{\tabletitle{Without post-processing}} \\
    SelfMask (coarse) \cite{shin2022selfmask} & & 81.1 & 40.3 & --- & 84.5 & 46.6 & --- & 89.3 & 64.6 & --- & --- & --- & --- \\
    LOST \cite{simeoni2021lost} & & 79.7 & 41.0 & 47.3 & 87.1 & 51.8 & 61.1 & 89.5 & 65.4 & 75.8 & 95.2 & 68.8 & 78.9 \\ 
    DSM \cite{melas2022deepsectralmethod} & & 80.8 & 42.8 & 55.3 & 84.1 & 47.1 & 62.1 & 86.4 & 64.5 & 78.5 & 94.1 & 66.7 & 82.9 \\ 
    MOST \cite{rambhatla2023most} &  & 87.0 & 47.5 & 57.0 & 89.7 & 53.8 & 66.6 & 89.0 & 63.1 & 79.1 & --- & --- & --- \\
    FOUND-coarse \cite{simeoni2023found} & & --- & --- & --- & --- & --- & --- & 90.6 & 70.9 & 78.0 & --- & --- & --- \\
    TokenCut \cite{wang2022tokencut} & & 88.0 & 53.3 & 60.0 & 90.3 & 57.6 & 67.2 & 91.8 & 71.2 & 80.3 & 96.4 & 74.8 & 82.1 \\ 
    \midrule
    \multicolumn{14}{c}{\tabletitle{With post-processing}} \\
    LOST \cite{simeoni2021lost} & BS & 81.8 & 48.9 & 57.8 & 88.7 & 57.2 & 69.7 & 91.6 & 72.3 & 83.7 & 96.6 & 77.6 & 84.3 \\ 
    DSM \cite{melas2022deepsectralmethod} & CRF & 87.1 & 56.7 & 64.4 & 83.8 & 51.4 & 56.7 & 89.1 & 73.3 & 80.5 & 96..6 & 76.9 & 84.3 \\
    FOUND-coarse \cite{simeoni2023found} & BS & --- & --- & --- & --- & --- & --- & 90.9 & 71.7 & 79.2 & --- & --- & --- \\
    TokenCut \cite{wang2022tokencut} & BS & \bf 89.7 & \bf 61.8 & \bf 69.7 & \bf 91.4 & 62.4 & \bf 75.5 & \bf 93.4 & 77.2 & \bf 87.4 & \bf 97.4 &\bf  79.5 & \bf 87.1 \\ 
    PaintSeg (on TokenCut)\cite{li2023paintseg} & \cite{li2023paintseg} & --- & --- & --- & --- & \bf 67.0 & --- & --- & \bf 80.6 & --- & --- & --- & --- \\ 
    \bottomrule
\end{tabular}
    }
    \caption{\label{tab:unsup_saliency_onlyss-feats} 
    \textbf{Evaluation of \unsupsaliency{} methods}. Comparisons of methods solely leveraging pre-trained self-supervised features (discussed in \autoref{sec:ssonly}). The task is described in \autoref{sec:background:def:unsup-saliency}. We distinguish the results obtained with and without post-processing.
    More results for this task, including the ones obtained with learning-based methods, are provided in \autoref{tab:unsup_saliency}.
    }
\end{table*}

\subsection{Quantitative results}
\label{sec:quantitative-onlyfeats}

We quantitatively compare the methods presented in this section, utilizing reported results from various papers\footnote{Note that, in certain methods that involve training, masks were generated as a preliminary step before the training phase. However, no results evaluating these preliminary masks were reported.}. Specifically, in \autoref{sec:quantitative-onlyfeats:single}, we discuss methods for unsupervised object discovery, while in \autoref{sec:quantitative-onlyfeats:saliency}, we focus on methods dedicated to unsupervised saliency detection. The comparison of class-agnostic multi-object detection or segmentation methods is deferred for subsequent discussion (\autoref{sec:sota}).

\begin{table}[ht!]
    \centering
    \rowcolors{2}{gray!15}{white}
\begin{tabular}{@{}l *{3}{c} @{}}
\toprule
    Method & VOC07 & VOC12 & CO20k \\
    \midrule
    \multicolumn{4}{c}{\tabletitle{SoTA before self-supervised ViTs era}} \\
    LOD \cite{vo2021largescale} & 53.6 & 55.1 & 48.5 \\
    rOSD \cite{Vo20rOSD} & 54.5 & 55.3 & 48.5 \\
    \midrule
    \multicolumn{4}{c}{\tabletitle{Leveraging self-supervised features}} \\
    DINO \cite{caron2021dino} & 45.8 & 46.2 & 42.1 \\
    LOST \cite{simeoni2021lost} & 61.9 & 64.0 & 50.7 \\ 
    DSS \cite{melas2022deepsectralmethod} & 62.7 & 66.4 & 52.2 \\
    TokenCut \cite{wang2022tokencut} & \bf 68.8 & \bf 72.1 & \bf 58.8 \\
\bottomrule
\end{tabular}
    \caption{\label{tab:single-obj-no-training} \textbf{Single-object discovery.} Results for the methods solely leveraging pre-trained self-supervised features (discussed in \autoref{sec:ssonly}).
    The task is described in \autoref{sec:background:def:single-object} and evaluated with the Correct Localization (CorLoc) metric.
    More results for this task, including the ones obtained with learning-based methods, are provided in \autoref{tab:single-obj}.}
\end{table}

\subsubsection{Single-object discovery}
\label{sec:quantitative-onlyfeats:single}

\OS{The capability of methods to perform \singleobjectdiscovery{} is typically evaluated using the CorLoc metric as detailed in \autoref{sec:background:def:single-object}.}
We report in \autoref{tab:single-obj-no-training} CorLoc results on the datasets VOC07 \cite{pascal-voc-2007}, VOC12 \cite{pascal-voc-2012}, COCO20k \cite{coco2014}. \OS{We report best results of each method obtained with the backbone model ViT-S/16 pre-trained following DINO~\cite{caron2021dino}. If interested, authors of TokenCut produce an interesting table (Tab. 5 in their paper) to compare different backbones.}

We observe that LOST~\cite{simeoni2021lost}, TokenCut~\cite{wang2022tokencut} and Deep Spectral Methods~\cite{melas2022deepsectralmethod} (noted DSS in the table) largely surpass previous baselines, and thus without performing dataset-level optimization. Also TokenCut obtains best results and produces a good prediction in $\approx70\%$ of the time on both VOC07 and VOC12. In the more challenging dataset COCO, methods are still able to predict an accurate box half of the time.
It is to be noted that TokenCut and DSS both require to compute eigenvectors which has some computational cost.

\subsubsection{\Unsupsaliency}
\label{sec:quantitative-onlyfeats:saliency}

The quality of the multi-object localization can be measured using the \unsupsaliency{} protocol (detailed in \autoref{sec:background:def:unsup-saliency})  which evaluates how well the foreground pixels are separated from the background ones.
We report results provided by the authors or following works in \autoref{tab:unsup_saliency_onlyss-feats}.

\parag{Comparison of methods.}
We first focus on the results of the methods without the application of post-processing. We can observe that best results are obtained by TokenCut \cite{wang2022tokencut} again on all datasets. Results of LOST and DSS are on par with one another, DSS being better on DUT-OMRON and LOST better on other datasets.
It is to be noted that methods noted with `coarse' are not the final results of the methods and are further improved by using training strategies discussed in \autoref{sec:training}. 
Also, MOST \cite{rambhatla2023most} is not designed directly for the task of \unsupsaliency{}; in order to produce a score the authors select the largest pool and use as saliency map the similarity map computed using its tokens.

\begin{figure}[ht!]
    \centering
    \includegraphics[width=0.95\columnwidth]{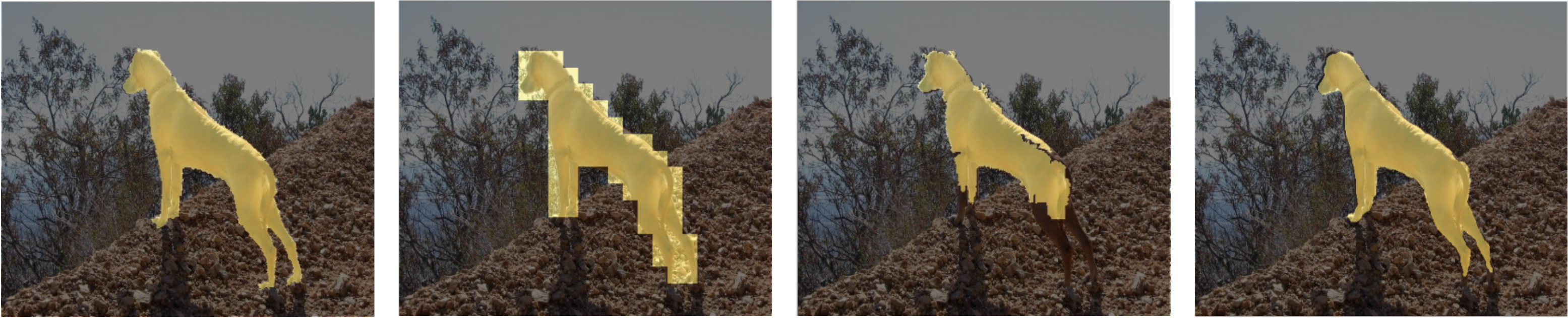}
    \footnotesize
    \noindent \textcolor{white}{louuu} GT \textcolor{white}{louuu} TC \cite{wang2022tokencut} \textcolor{white}{looou} TC+BS \textcolor{white}{louu} PaintSeg \cite{li2023paintseg} 
    \caption{\textbf{Visualization of the impact of the post-processing}. We observe improvements on the borders of the mask with the application of Bilateral Solver (BS) \cite{bilateralsolverbarron2016} and PaintSeg \cite{li2023paintseg} on the results of TokenCut \cite{wang2022tokencut}, noted `TC'. The figure is borrowed from PaintSeg \cite{li2023paintseg}.}
    \label{fig:post-process}
\end{figure}

\parag{Post-processing refinement.}
We now discuss the impact of the post-processing step and observe that in all cases, results are significantly boosted. For instance LOST, DSM and TokenCut are all boosted by around 8\,pts of IoU on DUT-OMRON showing the importance and opportunity provided by post-processing to refine masks. 
Moreover, by employing the latest post-processing strategy, PaintSeg~\cite{li2023paintseg}, the results of TokenCut are further improved.
We present the visual results of TokenCut with the Bilateral Solver (BS) and PaintSeg in \autoref{fig:post-process}; 
we observe that while the Bilateral Solver (BS) refines the mask, it concurrently introduces degradation, observed on the leg of the dog—an acknowledged phenomenon reported in previous studies \cite{simeoni2023found}.
In contrast, the output of PaintSeg aligns more closely with the ground truth.

\subsection{Limits}
\OS{Leveraging self-supervised features of transformers allow us} to \emph{discover objects with no annotation} and achieve interesting single-object discovery performances on VOC and COCO datasets. However, extending this capability to multi-object discovery is not straight-forward and several questions remain on how to:
\begin{itemize}
    \item successfully perform multi-object detection? 
    \item exchange information at a \emph{dataset level}?
    \item refine results? 
\end{itemize}

We examine in the next section how integrating a simple learning step can greatly boost results and generalizability properties.

%
\section{Training with coarse pseudo-labels}
\label{sec:training}

\begin{figure}[ht!]
    \centering
    \begin{subfigure}{\linewidth}
        \centering
        \includegraphics[width=0.7\linewidth]{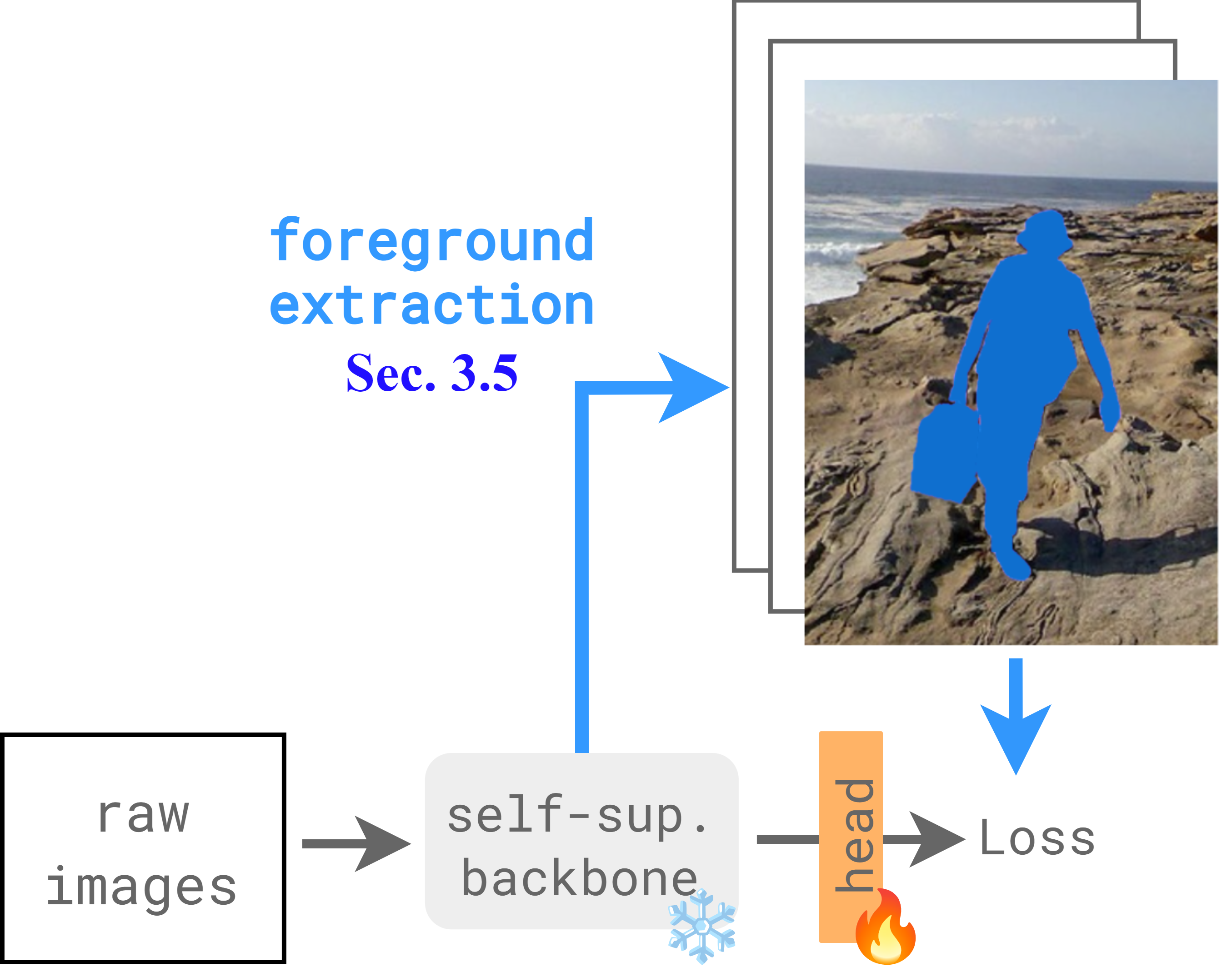}
    \caption{Training \emph{a head} to extract information from the self-supervised features (\autoref{sec:training:head}).}
    \label{fig:trainings:head}
    \end{subfigure}
    \begin{subfigure}{\linewidth}
        \includegraphics[width=\linewidth]{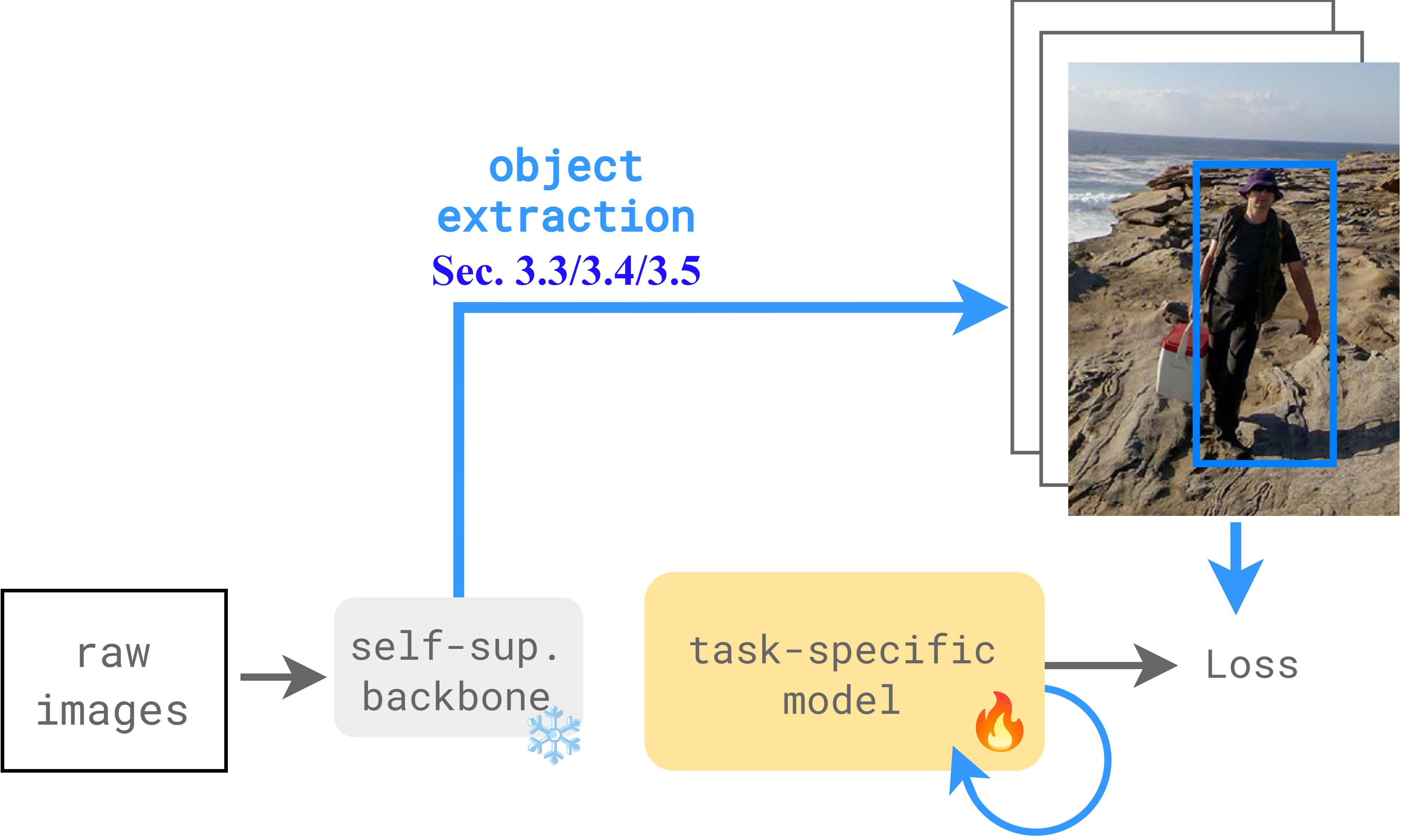}
    \caption{Training a task-specific model (\autoref{sec:training:task-specific}).}
    \label{fig:trainings:taskspec}
    \end{subfigure}
    \begin{subfigure}{\linewidth}
        \includegraphics[width=\linewidth]{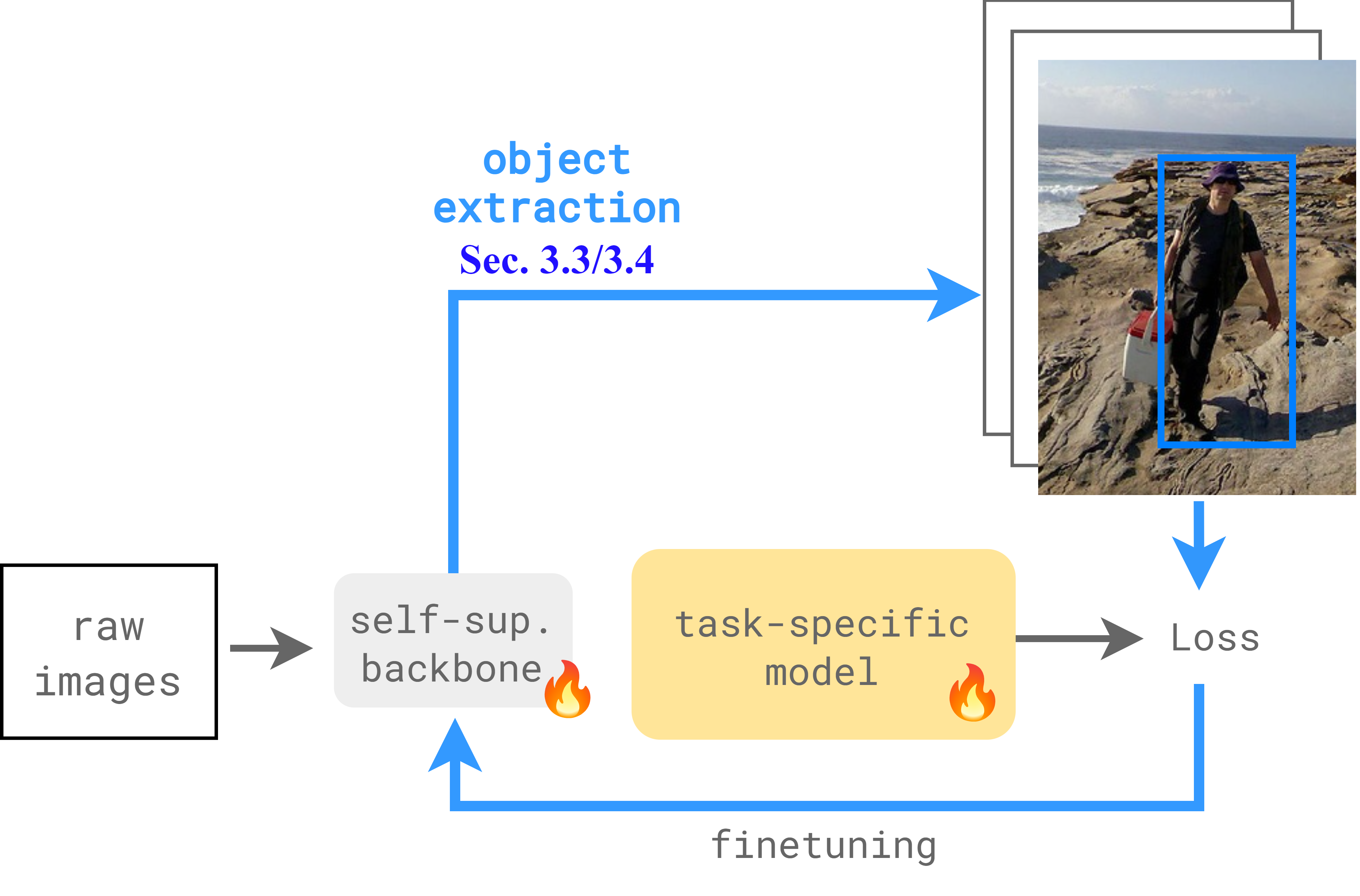}
    \caption{Improving the self-supervised features (\autoref{sec:training:finetuning}).}
    \label{fig:trainings:finetuning}
    \end{subfigure}
    \caption{
    Different unsupervised training strategies presented in \autoref{sec:training}. 
    By using the coarse localization predictions of training-free methods (presented in \autoref{sec:ssonly}) as pseudo-labels, 
    these unsupervised training methods can build new models with improved localization performance.}
    \label{fig:trainings}
\end{figure}

In \autoref{sec:ssonly}, we presented training-free methods designed to extract as much information as possible from features of an already pre-trained backbone with no additional training. Here we discuss unsupervised methods that incorporate training in order to further improve object localization performance. Indeed, by using the coarse localization predictions of training-free methods as \emph{pseudo-labels} \cite{simeoni2021lost, simeoni2023found, wang2023cutler, wang2022freesolo}, we can build new models that make better localization predictions than those used for their training. 
In particular, we discuss methods that use pseudo-labels: (a) for 
\emph{training prediction heads for foreground segmentation} on top of the frozen self-supervised features (\autoref{sec:training:head}), (b) for \emph{end-to-end training of task-specific architectures} using techniques for handling the noise in pseudo-labels (\autoref{sec:training:task-specific}), or (c) for \emph{unsupervised finetuning} of the self-supervised pre-trained backbones 
(\autoref{sec:training:finetuning}).
These three approaches are illustrated in \autoref{fig:trainings}.

\subsection{Training prediction heads on top of self-supervised features}
\label{sec:training:head}

We have seen in \autoref{sec:ssonly} that raw self-supervised features already contain sufficient localization properties to successfully extract foreground or objects with well-designed algorithms. Another type of methods \cite{simeoni2023found, ravindran2023sempart, bielski2022move, zhang2023ucos-da} propose to train a simple head on top of these raw features to perform similar extractions in a feedforward manner (see \autoref{fig:trainings:head}).
These methods focus on foreground segmentation, a task that necessitates pixel-wise binary predictions and which can be performed using simple models on top of powerful self-supervised features.

\parag{Linear separation of the features.}
One of the simplest ways to exploit DINO features is probably to add a \emph{single} 1x1 convolution layer on top of them \cite{simeoni2023found}. 
The method \OS{FOUND~\cite{simeoni2023found}} learns to project the self-supervised features to an `objectness' space and is trained using two objectives: 
the first 
guides the prediction towards the coarse masks \gilles{of \autoref{sec:found-coarse}}; 
\gilles{the second guides the predictions towards its own post-processed version. Specifically, the authors use as post-processing the Bilateral Solver operation which 
improves the quality of the masks and is discussed in \autoref{sec:post-processing}.}
This work shows that foreground and background patches are linearly separable in the feature space of a DINO-pretrained ViT.
Song et al. \cite{song2023UOLwRPS} \OS{also} employ a linear layer trained on frozen self-supervised pre-trained patch features \OS{but} using the Representer Point Selection \cite{yeh2018representer} framework. They utilize soft pseudo-masks derived from the magnitude of these pre-trained features as foreground-background target masks: higher feature magnitudes in patches \gilles{indicate} 
the presence of a salient object.

\parag{Towards higher-resolution predictions.}
Alternatively, SEMPART \cite{ravindran2023sempart} incorporates a transformer layer, followed by a 1x1 convolution layer with a sigmoid activation unit \gilles{on top of} the fixed DINO features. 
The training objective minimizes the normalized cut loss across a graph defined by an adjacency matrix as defined in \eqref{eq:sim}.
In other words, SEMPART is crafted to learn to predict the (soft) bi-partition of a graph that has similar properties as the one used in, e.g., LOST or TokenCut (see \autoref{sec:no-training}).
As the mask at the output of this first step is defined at patch-level and is thus coarse,
SEMPART jointly trains an upconvolution network to refine the prediction. This network processes and complements the features extracted from the transformer layer with RGB features at progressively increasing resolutions, resulting in a finer mask at the original image resolution. The training process for these refined masks involves a coarse-to-fine mask distillation process wherein the finer masks, after being downsampled through average pooling, are required to match the coarse masks.
Furthermore, SEMPART smooths the predictions and ensures the preservation of fine boundary details through the incorporation of graph-driven total-variation regularization losses.
In summary, instead of partitioning a graph to obtain a first mask and improving the quality of this mask by post-processing, SEMPART is able to learn this full process end-to-end by making each step differentiable. Let us mention that DeepCut \cite{aflalo2022deepcut} also minimizes the normalized cut loss (or some variation) to train a lightweight graph neural network on top of frozen DINO features.

\parag{Learning by moving objects.}
MOVE \cite{bielski2022move} exploits two self-supervised ViTs: the first with DINO; the second with MAE. A segmentation head is trained on top of the frozen DINO features with the help of the MAE-pretrained ViT. The training objective is constructed on the fact that if objects are accurately segmented and the background inpainted using the pretrained MAE, then the objects can be pasted at a different location in the image without generating duplication artifacts. The segmentation head is trained using an adversarial loss design to detect these duplication artifacts.
The concept that foreground objects can be ``realistically moved" within an image or to other images has also been investigated in different works e.g. \cite{remez2018learning, ostyakov2018seigan, bielski2019emergence, arandjelovic2019object, yang2019unsupervised,katircioglu2021self}.

\parag{Leveraging low-level information.} The idea of modifying DINO features is also exploited in UCOS-DA~\cite{zhang2023ucos-da} to address the problem of the discovery of camouflaged objects, i.e., when objects and background share similar colors and textures. 
UCOS-DA~\cite{zhang2023ucos-da} also trains an additional head on top of a frozen DINO encoder using pseudo-masks generated by a training-free \gilles{foreground/background discovery} method. 
It also incorporates a foreground-background contrastive objective to improve \gilles{the performance on this task}.

\subsection{Training a task-specific model}
\label{sec:training:task-specific}

Tasks such as object detection or instance segmentation are hard to handle by training a simple head on top of self-supervised features as in the previous section. Instead, they are easier way to address the tasks with dedicated network architectures. Several methods \cite{simeoni2021lost,shin2022selfmask,wang2023cutler,bielski2022move} have shown that these architectures can actually be trained using pseudo-labels derived from the coarse predictions of \autoref{sec:ssonly}. In \autoref{sec:training:task-specific:general}, we present how these models are trained and highlight their benefit.
We specifically discuss the problem of noise in pseudo-labels and solutions to mitigate it in \autoref{sec:training:task-specific:tricks}.

\subsubsection{Training principle and benefits}
\label{sec:training:task-specific:general}

\parag{General idea.} Given the task of interest, the idea is to use a standard task-specific model which will be trained using as ground-truth the \emph{pseudo-labels} generated with the training-free unsupervised methods described in \autoref{sec:ssonly}. %
This strategy is illustrated in \autoref{fig:trainings:taskspec}. Note that the pseudo-labels do not have any semantic information %
and that the task-specific models are therefore trained %
in class-agnostic fashion.

\parag{Task-specific models.}
Focusing on the object detection task, the authors of LOST \cite{simeoni2021lost} propose to train the popular object detector Faster R-CNN~\cite{ren2015fasterrcnn} (without adaptation) with LOST single-object discovery predictions (\autoref{eq:lost}) as pseudo-labels. In order to generate instance-level masks, FreeSOLO \cite{wang2022freesolo} utilizes the pseudo instance-masks generated thanks to the FreeMask methodology (detailed in \autoref{sec:3:single-to-multi:adaptive}), for training a SOLO-based \cite{wang2020solo} instance segmentation model. Subsequent exemplar-FreeSOLO \cite{ishtiak2023examplar-freesolo} enhances FreeSOLO by introducing an unsupervised mechanism for generating \emph{object exemplars} \OS{which can be seen as a vocabulary} and employ a contrastive loss to enhance the discriminative capability of the SOLO-based instance segmenter.
\OS{Alternatively, CutLer \cite{wang2023cutler} and IMST \cite{lim2022imst}} train a class-agnostic Mask R-CNN~\cite{he2017mask} instance segmentation model.
\OS{CutLer model is trained with}
initial coarse instance masks \OS{generated} for each image via their MaskCut methodology (detailed in \autoref{sec:3:single-to-multi:iterative}). 
For the task of foreground/background segmentation
SelfMask \cite{shin2022selfmask} and MOVE \cite{bielski2022move}
train a dedicated segmenter network (a variant of MaskFormer~\cite{cheng2021maskformer}).

\parag{Benefits.} The methods \cite{simeoni2021lost,shin2022selfmask,wang2023cutler,bielski2022move,lim2022imst,wang2022freesolo, ishtiak2023examplar-freesolo} have shown that training a task-specific model using unsupervised coarse predictions as pseudo-labels permits us to (1) obtain predictions adapted to the task, (2) improve the overall precision of the predictions, (3) go from single object to multi-object detection and (4) increase the prediction recall. 
For instance, the authors of LOST~\cite{simeoni2021lost} have shown that training a Faster R-CNN, called Class-Agnostic Detector (CAD) in this context, with such pseudo-labels leads to detection of better quality, as seen with the improvement of the CorLoc scores in \autoref{tab:res-CAD}. Training this Class-Agnostic Detector over a (relatively) large dataset offers a regularization effect over all pseudo-labels, which improves the quality of the predictions.
Additionally, this CAD is able to output several boxes per image 
even if trained with a single pseudo-box per image.

\begin{table}[t]
    \centering
    \resizebox{\linewidth}{!}{
    \begin{tabular}{lc|ccc}
        Method & + CAD & VOC07 & VOC12 & COCO20k \\
        \toprule
       \rowcolor{PineGreen!10!white}  LOST~\cite{simeoni2021lost}  & & 61.9 & 64.0 & 50.7 \\
         & \checkmark & 65.7 \improv{3.8} & 70.4 \improv{6.4} & 57.5 \improv{6.8} \\
       \rowcolor{PineGreen!10!white}  TokenCut~\cite{wang2022tokencut} & & 68.8 & 72.1 & 58.8 \\
         & \checkmark & 71.4 \improv{2.6} & 75.3 \improv{3.2} & 62.6 \improv{3.8}\\
       \rowcolor{PineGreen!10!white} MOVE~\cite{bielski2022move} & & 76.0 & 78.8 & 66.6 \\
          & \checkmark & 77.1  \improv{1.1} & 80.3 \improv{1.5} & 69.1 \improv{2.5} \\
         
    \end{tabular}
    }
    \caption{\small{\emph{Multi-object discovery} results using the +CAD strategy on the VOC and COCO datasets for different \emph{self-supervised feature extraction} methods \cite{simeoni2021lost, wang2022tokencut, bielski2022move} \colorbox{PineGreen!10!white}{highlighted} in the table. Evaluation with the corloc metric.}}
    \label{tab:res-CAD}
\end{table}

\subsubsection{Dealing with the noise in the pseudo-labels}
\label{sec:training:task-specific:tricks}

\noindent Task-specific models, like object detectors \cite{ren2015fasterrcnn} and instance-segmentation models \cite{he2017mask,cheng2021maskformer,wang2020solo}, have been conceived to generate the desired predictions for a certain task. However they have been designed to be trained in a fully-supervised fashion, with `perfect' ground-truth annotations. In our case, the pseudo-labels used are noisy and might not cover all objects in an image. In order to deal with such difficulty, a set of methods propose adaptation of the training losses or of the training scheme \cite{wang2023cutler, wang2022freesolo}.

\parag{Dealing with noisy predictions.} To enhance stability in the presence of inherently noisy pseudo-masks, FreeSOLO introduces modifications to the training protocol of SOLO.
Drawing inspiration from the weakly-supervised literature, they project SOLO-predicted instance masks and pseudo masks onto the x and y axes, enforcing alignment through a corresponding loss. At the same time, 
they incorporate a pairwise affinity loss leveraging the expectation that proximal pixels with similar color should share the same class in predicted masks (foreground or background class).

In CutLer \cite{wang2023cutler}, the authors enhance the training of Mask R-CNN by employing a loss dropping strategy which excludes regions not covered by
any pseudo-mask. This strategy \emph{boosts robustness} to objects overlooked by the pseudo-labels.

\parag{Improving through several training phases.} \OS{In an attempt to further augment the number of boxes produced per image, CutLer~\cite{wang2023cutler} proposes to re-train the detection model 
several times, each round using the coarse boxes or the prediction of the previous training. In practice, they perform three such rounds and show that the number of produced detection increases after each round.}

\parag{Learning to denoise and merge object parts with classifiers.}
We have seen in \autoref{sec:3:single-to-multi:iterative} that UMOD \cite{kara2023umod} leverages iterative clustering with a stopping criterion which allows the estimation of the number of objects in an image. This process is nevertheless imperfect resulting in discovery of object parts rather than complete objects. The authors therefore post-process the masks by merging object parts using two convolutional classifiers trained with the help of automatically extracted pseudo-labels. The first is trained to distinguish foreground/object parts from background parts. The second applies on foreground parts to assign them a pseudo-class representing semantic concepts discovered at the level of the dataset.

\vspace{2em}
\subsection{Unsupervised ViT fine-tuning} 
\label{sec:training:finetuning}

We now discuss methods 
\PP{that} finetune directly DINO-pretrained ViT backbones (see \autoref{fig:trainings:finetuning}) without loosing any of their object localization properties and actually improving their quality for the tasks of interest.

\begin{table*}[ht!]
    \setlength{\tabcolsep}{2.8pt}
    \centering
    \resizebox{\linewidth}{!}{
    \rowcolors{2}{gray!15}{white}
    \begin{tabular}{l|c|ccc|ccc|ccc|ccc}
        \toprule
        & post- & \multicolumn{3}{c}{DUT-OMRON} & \multicolumn{3}{c}{DUTS-TE} & \multicolumn{3}{c}{ECSSD} & \multicolumn{3}{c}{CUB} \\ 
        Method & proc. & Acc & IoU & m$F_\beta$ & Acc & IoU & m$F_\beta$ & Acc & IoU & m$F_\beta$ & Acc & IoU & m$F_\beta$ \\ 
        \midrule 
        \multicolumn{14}{c}{\tabletitle{SoTA before self-supervised ViTs era}} \\
          HS \cite{yan2013hs} &  & 84.3 & 43.3 & 56.1 & 82.6 & 36.9 & 50.4 & 84.7 & 50.8 & 67.3 & --- & --- & ---\\
         wCtr \cite{zhu2014wctr} & & 83.8 & 41.6 & 54.1 & 83.5 & 39.2 & 52.2 & 86.2 & 51.7 & 68.4 & --- & --- & ---\\
         WSC \cite{li2015wsc}  & & 86.5 & 38.7 & 52.3 & 86.2 & 38.4 & 52.8 & 85.2 & 49.8 & 68.3 & --- & --- & ---\\
         DeepUSPS \cite{nguyen2019deepusps} & &  77.9 & 30.5 & 41.4 & 77.3 & 30.5 & 42.5 & 79.5 & 44.0 & 58.4 & --- & --- & ---\\
         BigBiGAN \cite{voynov2021biggan}  &  & 85.6 & 45.3 & 54.9 & 87.8 & 49.8 & 60.8 & 89.9 & 67.2 & 78.2 & --- & --- & ---\\
         E-BigBiGAN \cite{voynov2021biggan}  & & 86.0 & 46.4 & 56.3 & 88.2 & 51.1 & 62.4 & 90.6 & 68.4 & 79.7 & --- & --- & ---\\
         Melas-Kyriazi et al. \cite{melas2021}  & & 88.3 & 50.9 & --- & 89.3 & 52.8 & --- & 91.5 & 71.3 & --- & --- & --- & ---\\
         
        \midrule 
        \multicolumn{14}{c}{\tabletitle{Without post-processing}} \\
        MOST \cite{rambhatla2023most} & & 87.0 & 47.5 & 57.0 & 89.7 & 53.8 & 66.6 & 89.0 & 63.1 & 79.1 & --- & --- & --- \\
        WSCUOD \cite{lv2023wscuod} & & 89.7 & 53.6 & 64.4 & 91.7 & 89.9 & 73.1 & 92.2 & 72.7 & 85.4 & 96.8 & 77.8 & 87.9 \\ 
        FreeSOLO \cite{wang2022freesolo} & & 90.9 & 56.0 & 68.4 & 92.4 & 61.3 & 75.0 & 91.7 & 70.3 & 85.8 & --- & --- & --- \\ 
        LOST \cite{simeoni2021lost} & & 79.7 & 41.0 & 47.3 & 87.1 & 51.8 & 61.1 & 89.5 & 65.4 & 75.8 & 95.2 & 68.8 & 78.9 \\  
        DSM \cite{melas2022deepsectralmethod} & & 80.8 & 42.8 & 55.3 & 84.1 & 47.1 & 62.1 & 86.4 & 64.5 & 78.5 & 94.1 & 66.7 & 82.9 \\
        TokenCut \cite{wang2022tokencut} & & 88.0 & 53.3 & 60.0 & 90.3 & 57.6 & 67.2 & 91.8 & 71.2 & 80.3 & 96.4 & 74.8 & 82.1 \\
        DeepCut: CC loss \cite{aflalo2022deepcut} & & --- & --- & --- & --- & 56.0 & --- & --- & 73.4 & --- & --- & 77.7 & --- \\ 
        DeepCut: N-cut loss \cite{aflalo2022deepcut} & &  --- & --- & --- & --- & 59.5 & --- & --- & 74.6 & --- & --- & 78.2 & --- \\  
        SelfMask \cite{shin2022selfmask} & & 90.1 & 58.2 & --- & 92.3 & 62.6 & --- & 94.4 & 78.1 & --- & --- & --- & --- \\ 
        FOUND \cite{simeoni2023found}-single & & 92.0 & 58.6 & 67.3 & 93.9 & 63.7 & 73.3 & 91.2 & 79.3 & 94.6 & --- & --- & --- \\ 
        FOUND-multi \cite{simeoni2023found} & & 91.2 & 57.8 & 66.3 & 93.8 & 64.5 & 91.5 & 94.9 & 80.7 & 95.5 & --- & --- & --- \\ 

        MOVE \cite{bielski2022move} & & 92.3 & 61.5 & 71.2 & 95.0 & 71.3 & 81.5 & 95.4 & 83.0 & 91.6 & --- & \bf 85.8 & --- \\ 
        UCOS-DA \cite{zhang2023ucos-da} & & --- & --- & --- & --- & --- & --- & 95.1 & 81.6 & 89.1 & --- & --- & --- \\ 
        SEMPART-fine \cite{ravindran2023sempart} & & 93.2 & 66.8 & 76.4 & \bf  95.9 & \bf 74.9 & 86.7 & \bf 96.4 & \bf 85.5 & \bf 94.7 & --- & --- & --- \\
        SelfMask \cite{shin2022selfmask} on MOVE \cite{bielski2022move} \textcolor{Purple}{\cite{ravindran2023sempart}} & & 93.3 & 66.6 & 75.6 & 95.4 & 72.8 & 82.9 & 95.6 & 83.5 & 92.1 & --- & --- & --- \\
        SelfMask \cite{shin2022selfmask} on SEMPART-fine \cite{ravindran2023sempart} & & \bf 94.2& \bf 69.8 & \bf 79.9  & 95.8 & \bf 74.9 & \bf 87.9 & 96.3 & 85.0  &94.4 & --- & --- & --- \\
        
        \midrule
        \multicolumn{14}{c}{\tabletitle{With post-processing}} \\
        LOST \cite{simeoni2021lost} & BS & 81.8 & 48.9 & 57.8 & 88.7 & 57.2 & 69.7 & 91.6 & 72.3 & 83.7 & 96.6 & 77.6 & 84.3 \\
        DSM \cite{melas2022deepsectralmethod} & CRF & 87.1 & 56.7 & 64.4 & 83.8 & 51.4 & 56.7 & 89.1 & 73.3 & 80.5 & 96.6 & 76.9 & 84.3 \\ 
        WSCUOD \cite{lv2023wscuod} & BS & 90.9 & 58.5 & 68.3 & 92.5 & 63.0 & 76.4 & 92.8 & 74.2 & 89.6 & 97.3 & 79.7 & 89.3 \\ 
        TokenCut \cite{wang2022tokencut} & BS & 89.7 & 61.8 & 69.7 & 91.4 & 62.4 & 75.5 & 93.4 & 77.2 & 87.4 & 97.4 & 79.5 & 87.1 \\ 
        FOUND-single \cite{simeoni2023found} & BS & 92.1 & 60.8 & 70.6 & 94.1 & 64.5 & 76.0 & 94.9 & 80.5 & 93.4 & --- & --- & --- \\  
        FOUND-multi \cite{simeoni2023found} & BS & 92.2 & 61.3 & 70.8 & 94.2 & 66.3 & 76.3 & 95.1 & 81.3 & 93.5 & --- & --- & --- \\ 
        SelfMask \cite{shin2022selfmask} & BS & 91.9 & 65.5 & --- & 93.3 & 66.0 & --- & 95.5 & 81.8 & --- & --- & --- & --- \\ 
        PaintSeg (on TokenCut)\cite{li2023paintseg} & \cite{li2023paintseg} & --- & --- & --- & --- & 67.0 & --- & --- & 80.6 & --- & --- & --- & --- \\ 
        MOVE \cite{bielski2022move} & BS & 93.1 & 63.6 & 73.4 & 95.1 & 68.7 & 82.1 & 95.3 & 80.1 & 91.6 & --- & --- & --- \\ 
        SelfMask \cite{shin2022selfmask} on MOVE \cite{bielski2022move}\textcolor{Purple}{\cite{ravindran2023sempart}} & BS & 93.7 & 66.5 & 76.6 & 95.2 & 68.7 & 82.7 & 95.2 & 80.0 & 91.7 & --- & --- & --- \\

        \bottomrule
    \end{tabular}
    }
    \caption{\label{tab:unsup_saliency} \textbf{\Unsupsaliency{} evaluation.} We reproduce here published results on the task described in \autoref{sec:background:def:unsup-saliency}. We note with a \textcolor{Purple}{purple} citation in which paper the score was found (if not from the original paper).  We produce results with and without a post-processing step in different table sections. We note the application of a post-processing step in the column `post-proc.' with `BS' (Bilateral Solver \cite{bilateralsolverbarron2016}), `CRF' ( \cite{krahenbuhl2011crf}) and PaintSeg \cite{li2023paintseg} is a post-processing strategy. 
    }
\end{table*}

\parag{Detect to finetune.}
The authors of \cite{gomel2023boxbasedrefinement} propose to refine self-supervised features for the task of single-object discovery by: (1) training a detection head directly applied on top of a frozen self-supervised backbone, (2) freezing this detection head and (3) finetuning the self-supervised backbone. The detection head is trained using boxes extracted, e.g., using LOST, TokenCut or MOVE. Once this detection head is trained, a fixed number of boxes is extracted per image to serve as supervision signal to finetune the self-supervised backbone. During finetuning, a regularization term is also added on the output features of the self-supervised backbone to prevent it to diverge too far from its original state. The newly finetuned features can then be re-used in, e.g., LOST, TokenCut, or MOVE for single-object discovery (see \autoref{tab:single-obj}).

\parag{Learning scene-centric features.}
\OS{Following the classic self-supervised setup,} DINO \OS{backbones}
\cite{caron2021dino} are trained on object-centric images from ImageNet \cite{jia2009imagenet}. When used on natural scene-centric images or images with complex (e.g., cluttered) background, Lv et al.\ \cite{lv2023wscuod} observe that DINO features are noisy. %
A promising solution explored in WSCUOD~\cite{lv2023wscuod} to solve this issue, is to finetune the self-supervised features on \emph{scene-centric} images, e.g., DUTS \cite{wang2017duts-te}.
\OS{In order to} guide the finetuning on 
such images 
and encourage the suppression of the background activation, the authors of \cite{lv2023wscuod} propose two strategies.
The first is a
weakly-supervised contrastive learning (WCL) \cite{zheng2021wcl} to explore inter-image semantic relationships. WCL assigns weak labels to images based on graph-based similarity and then enhances image similarity through supervised contrastive learning.
The second is
a pixel-level semantic alignment loss \cite{xiao2022semantic_correspondence} that encourages the pixel-level consistency of the same object across different views of the same image.

\setlength{\tabcolsep}{1.5pt}
\begin{table}[t]
    \centering
    \resizebox{\linewidth}{!}{
    \rowcolors{2}{gray!15}{white}
    \begin{tabular}{l|ccc}
    \toprule
        Method & VOC07 & VOC12 & CO20k \\
    \midrule
    
        \multicolumn{4}{c}{\tabletitle{SoTA before self-supervised ViTs era}} \\
        Selective Search \cite{uijlings2013selectivesearch} & 18.8 & 20.9 & 16.0 \\
        EdgeBoxes \cite{zitnick2014edgebox} & 31.1 & 31.6 &  28.8 \\
        Kim et al. \cite{kim2009unsup_detection} & 43.9 & 46.4 & 35.1 \\
        Zhang et al. \cite{zhang2020object} & 46.2 & 50.5 &  34.8 \\
        DDT+ \cite{Wei2019ddtplus} & 50.2 & 53.1 & 38.2 \\ 
        rOSD \cite{vo2020unsup_multi_object_discovery} & 54.5 & 55.3 & 48.5 \\
        LOD \cite{vo2021largescale} & 53.6 &  55.1 & 48.5 \\
        
        \midrule
        \multicolumn{4}{c}{\tabletitle{No learning}} \\
        
        DINO \cite{caron2021dino} & 45.8 & 46.2 & 42.1 \\
        LOST \cite{simeoni2021lost} & 61.9 & 64.0 & 50.7 \\ 
        DSS \cite{melas2022deepsectralmethod} & 62.7 & 66.4 & 52.2 \\
        TokenCut \cite{wang2022tokencut} & 68.8 & 72.1 & 58.8 \\ 
        
        \midrule

        \multicolumn{4}{c}{\tabletitle{With learning}} \\
        FreeSOLO \cite{wang2022freesolo}\textcolor{Purple}{\cite{bielski2022move}} & 56.1 & 56.7 & 52.8 \\
        LOST \cite{simeoni2021lost} + CAD  & 65.7 & 70.4 & 57.5 \\ 
        DeepCut: CC loss \cite{aflalo2022deepcut} & 68.8 & 67.9 & 57.6 \\
        DeepCut: N-cut loss \cite{aflalo2022deepcut} & 69.8 & 72.2 & 61.6 \\
        WSCUOD \cite{lv2023wscuod} & 70.6 & 72.1 & 63.5 \\
        TokenCut \cite{wang2022tokencut} + CAD  & 71.4 & 75.3 & 62.6 \\ 
        SelfMask \cite{shin2022selfmask} & 72.3 & 75.3 & 62.7 \\
        FOUND \cite{simeoni2023found} & 72.5 & 76.1 & 62.9 \\
        SEMPART-Coarse \cite{ravindran2023sempart} & 74.7 & 77.4 & 66.9 \\ 
        SEMPART-Fine \cite{ravindran2023sempart} & 75.1 & 76.8 & 66.4 \\
        IMST \cite{lim2022imst} & 76.9 & 78.7 & \bf 72.2 \\
        MOVE \cite{bielski2022move} & 76.0 & 78.8 & 66.6 \\
        MOVE \cite{bielski2022move} + CAD & 77.1 & 80.3 & 69.1 \\
        MOVE multi \cite{bielski2022move} + CAD & 77.5 & \bf 81.5 & 71.9 \\
        Bb refin. \cite{gomel2023boxbasedrefinement} w. MOVE & 77.5 & 79.6 & 67.2 \\
         Bb refin. \cite{gomel2023boxbasedrefinement} w. MOVE + CAD & \bf 78.7 & 81.3 & 69.3 \\
    \bottomrule
    \end{tabular}
    }
    \caption{\label{tab:single-obj} \textbf{Single-object discovery.} The task is described in \autoref{sec:background:def:single-object} and evaluated with the Correct Localation (CorLoc) metric on the datasets  VOC07 \cite{pascal-voc-2007}, VOC12 \cite{pascal-voc-2012} and COCO 20k \cite{coco2014,vo2021large_scale_unsup_object_discovery} (noted `CO20k'). If the score was not reported in the original paper, we cite in \textcolor{Purple}{purple} the paper which produced it.}
\end{table}

\subsection{State-of-the-art results}
\label{sec:sota}

We discuss in this section the state-of-art results obtained on the different tasks of unsupervised object localization (presented in \autoref{sec:background}). 
We report results published by the authors or following works for the tasks of: \unsupsaliency{}, \singleobjectdiscovery{}, \multiclassif{}, and \multiseg{}. Please refer to \autoref{sec:background} for the definition of the metrics and datasets used here.

\subsubsection{\Unsupsaliency{}}

\OS{We first evaluate the unsupervised methods presented in \autoref{sec:ssonly} and \autoref{sec:training} on the task of \unsupsaliency{}. This task requires to separate pixels of background from those of foreground objects and is detailed in \autoref{sec:background:def:unsup-saliency}. The scores are gathered in \autoref{tab:unsup_saliency} and include results with and without post-processing step.}

\OS{We can observe that best results are obtained with the recent SEMPART~\cite{ravindran2023sempart} \gilles{even when compared to methods that exploit post-processing,}
showing the interest \gilles{of learning to} produce high-resolution predictions. Moreover, training \gilles{the SelfMask auto-encoder \cite{shin2022selfmask}} 
(described in \autoref{sec:3:all-multi}) with the outputs of MOVE~\cite{bielski2022move} or SEMPART~\cite{ravindran2023sempart} as pseudo-masks boosts results in 
\gilles{either case} with up to $5$ pts of IoU.}

Finally, as discussed in detail in \autoref{sec:quantitative-onlyfeats:saliency}, the application of post-processing refinement permits obtaining more accurate outputs.

\subsubsection{\Singleobjectdiscovery{}}

\OS{First, we evaluate the capacity of the methods to produce a well-localized detection on \gilles{at least} one of the objects of interest. We gather all results in \autoref{tab:single-obj} and compare methods that integrate or not a learning step. 
Naturally, we observe that methods \gilles{that exploits a learning step through the dataset of interest} 
achieve higher scores than those solely exploiting \gilles{pre-trained} features. 
\gilles{The leaderboard is dominated by MOVE \cite{bielski2022move}, which can be improved by learning a class-agnostic detector (+CAD), and \cite{gomel2023boxbasedrefinement}, which improves upon MOVE results.}
Overall, several methods \cite{bielski2022move, lim2022imst, ravindran2023sempart, gomel2023boxbasedrefinement, simeoni2023found} achieve above $75$ pts of CorLoc on VOC07 and VOC12 which is an impressive result for methods using zero manual annotation. On the more challenging COCO dataset, which contains more and smaller objects, best results \cite{bielski2022move, lim2022imst} surpass $70$ pts of CorLoc.} 
\gilles{Let us mention that we did not include the score of MOST in \autoref{tab:single-obj} as the CorLoc reported in \cite{rambhatla2023most} is computed differently than in the related works}.

\begin{table*}[ht]
    \centering
    \resizebox{\linewidth}{!}{
    \rowcolors{2}{gray!15}{white}
    \begin{tabular}{@{} l *{17}{c} @{}}
        \toprule
        & ~ & \multicolumn{5}{c}{COCO20K} & \multicolumn{5}{c}{VOC07} & \multicolumn{3}{c}{VOC12} & \multicolumn{3}{c}{COCO val2017} \\
        
        \cmidrule(lr){3-7}\cmidrule(lr){8-12}\cmidrule(lr){13-15} \cmidrule(lr){16-18}
        Method & Detector & AP$_{50}$ & AP$_{75}$ & AP & odAP$_{50}$ & odAP & AP$_{50}$ & AP$_{75}$ & AP & odAP$_{50}$ & odAP & AP$_{50}$ & odAP$_{50}$ & odAP & AP$_{50}$ & AP$_{75}$ & AP \\ 
        \midrule
        \multicolumn{18}{c}{\tabletitle{No learning}} \\
        rOSD \cite{Vo20rOSD} & --- & --- & --- & --- & 5.2 & 1.6 & ~ & --- & --- & 13.1 & 4.3 & --- & 15.4 & 5.27 & --- & --- & --- \\
        LOD \cite{vo2021largescale} & --- & --- & --- & --- & 6.6 & 2.0 & --- & --- & --- & 13.9 & 4.5 & --- & 16.1 & 5.34 & --- & --- & --- \\
        TokenCut \cite{wang2022tokencut} & --- &  --- & --- & --- & --- & --- & --- & --- & --- & --- & --- & --- & --- & --- & 5.8 & 2.8 & 3.0 \\
        MOST \cite{rambhatla2023most} & --- & --- & --- & --- & --- & 1.7 & --- & --- & --- & --- & 6.4 & --- & --- & --- & --- & --- & \\
        
        \midrule
        \multicolumn{18}{c}{\tabletitle{With learning}} \\
        rOSD \cite{Vo20rOSD} + CAD  & FRCNN & 8.4 & --- & --- & --- & --- & 24.2 & --- & --- & --- & --- & 29.0 & --- & --- & --- & --- & --- \\
        LOD \cite{vo2021largescale} + CAD & FRCNN & 8.8 & --- & --- & 7.3 & 2.3 & 22.7 & --- & --- & 15.8 & 5.0 & 28.4 & 20.9 & 7.07 & --- & --- & --- \\
        LOST \cite{simeoni2021lost} + CAD & FRCNN & 9.9 & --- & --- & 7.9 & 2.5 & 29.0 & --- & --- & 19.8 & 6.7 & 33.5 & 24.9 & 8.85 & --- & --- & --- \\
        TokenCut \cite{wang2022tokencut} + CAD & FRCNN & 10.5 & --- & --- & --- & --- & 26.2 & ~ & ~ & --- & --- & 35.0 & --- & --- & --- & --- & ~ \\
        UMOD \cite{kara2023umod} & ResNet50 & 13.8 & --- & --- & 5.4 & 2.1 & 27.9 & --- & --- & 15.4 & 6.8 & 36.2 & ~ & --- & --- & --- & --- \\
        FreeSOLO \cite{wang2022freesolo} & SOLO & 12.4 & 4.4 & 5.6 & --- & --- & 24.5 & 7.2 & 10.2 & --- & --- & --- & --- & --- & 12.2 & 4.2 & 5.5 \\
        Ex.-FreeSOLO \cite{ishtiak2023examplar-freesolo} & SOLO & --- & --- & --- & --- & --- & 26.8 & 8.2 & 12.6 & --- & --- & --- & --- & --- & 17.9 & 8.6 & 12.6 \\
        WSCUOD \cite{lv2023wscuod} & FRCNN & 13.6 & --- & --- & --- & --- & 30.5 & --- & --- & --- & --- & --- & --- & --- & --- & --- & --- \\
        IMST \cite{lim2022imst} & MRCNN & --- & --- & --- & --- & --- & --- & --- & --- & --- & --- & --- & --- & --- & 18.1 & 7.6 & 8.8 \\
        MOVE \cite{bielski2022move} & ~ & ~ & --- & --- & --- & --- & --- & --- & --- & --- & --- & --- & --- & --- &  19 & 6.5 & 8.2 \\
        CutLER \cite{wang2023cutler} & MRCNN & 21.8 & 11.1 & 10.1 & --- & --- & --- & --- & --- & --- & --- & --- & --- & --- & 21.3 & 11.1 & 10.2 \\
        CutLER \cite{wang2023cutler} & C-MRCNN & 22.4 & 12.5 & 11.9 & --- & --- & 36.9 & 19.2 & 20.2 & --- & --- & --- & --- & --- & 21.9 & 11.8 & 12.3 \\ 
        \bottomrule
    \end{tabular}
    }
    \caption{\label{tab:multi-det}\textbf{\Multiclassif{}.} The task and metrics are described in \autoref{sec:background:def:multi-object-classif}. The task is described \autoref{sec:background:def:multi-object-classif}. We note the detector model used in the column `Detector': `FRCNN' stands for Faster R-CNN~\cite{ren2015fasterrcnn}, `MRCNN' for Mask R-CNN~\cite{he2017mask}, `C-MRCNN' for Cascade Mask R-CNN~\cite{cai2018cascade} and SOLO model is described in \cite{wang2020solo}. UMOD learns two classifiers which are based on a ResNet50~\cite{he2016residual} backbone. Notation `+CAD' denote training a Faster R-CNN \cite{ren2015fasterrcnn} following \cite{simeoni2021lost}.
    \vspace{15pt}
    }
\end{table*}

\begin{table*}[ht]
    \centering
    \resizebox{\linewidth}{!}{
    \rowcolors{2}{gray!15}{white}
    \begin{tabular}{@{} l *{14}{c} @{}}
    \toprule
        & & \multicolumn{3}{c}{COCO 20K \cite{coco2014,vo2020unsup_multi_object_discovery}} & \multicolumn{3}{c}{COCO val2017 \cite{coco2014}} & \multicolumn{3}{c}{PASCAL VOC12 \cite{pascal-voc-2012}} & \multicolumn{3}{c}{UVO val \cite{wang2021uvo}} \\ \cmidrule(lr){3-5}\cmidrule(lr){6-8}\cmidrule(lr){9-11} \cmidrule(lr){12-14}
        Method & Detector & AP$_{50}$ & AP$_{75}$ & AP & AP$_{50}$ & AP$_{75}$ & AP & AP$_{50}$ & AP$_{75}$ & AP  & AP$_{50}$ & AP$_{75}$ & AP \\ 
    \midrule
        \multicolumn{14}{c}{\tabletitle{No learning}} \\
        DINO \cite{caron2021dino} & --- & 1.7 & 0.1 & 0.3 & --- & --- & --- & 6.7 & 0.6 & 1.9 & --- & --- & --- \\ 
        TokenCut \cite{wang2022tokencut}  & --- & --- & --- & --- & 4.8 & 1.9 & 2.4 & --- & --- & --- & --- & --- & --- \\ 
        MaskDistill \cite{gansbeke2022maskdistill} (coarse) \textcolor{Purple}{\cite{lim2022imst}} & ---  & 3.1 & 0.5 & 1.3 & --- & --- & --- & 12.8 & 0.3 & 4.8  & --- & --- & --- \\ 
        IMST \cite{lim2022imst} (coarse) & --- & 5.6 & 1.2 & 2.1 & 4.6 & 1.0 & 1.7 & --- & --- & ---  & --- & --- & --- \\ 
        \midrule
        \multicolumn{14}{c}{\tabletitle{With learning}} \\
        MaskDistill \cite{gansbeke2022maskdistill} & MRCNN & 6.8 & 2.1 & 2.9 & --- & --- & --- & 24.3 & 6.9 & 9.9  & --- & --- & --- \\  
        IMST \cite{lim2022imst} & MRCNN & 15.4 & 5.6 & 6.9 & 14.8 & 5.2 & 6.6 & --- & --- & ---  & --- & --- & --- \\ 
        FreeSOLO \cite{wang2022freesolo} & SOLO & --- & --- & --- & 9.8 & 2.9 & 4.0 & --- & --- & --- & 12.7 & 3.0 & 4.8 \\ 
        Ex.-FreeSOLO \cite{ishtiak2023examplar-freesolo} & SOLO &  --- & --- & --- & 13.2 & 6.3 & 8.4 & --- & --- & --- & 14.2 & 7.3 & 9.2 \\ 
        CutLER \cite{wang2023cutler} & MRCNN & 18.6 & 9.0 & 8.0 & 18.0 & 8.9 & 7.9 & --- & --- & ---  & --- & --- & --- \\ 
        CutLER \cite{wang2023cutler} & C-MRCNN & 19.6 & 10 & 9.2 & 18.9 & 9.7 & 9.2 & --- & --- & --- & 22.8 & 8.0 & 10.1 \\ 
    \bottomrule
    \end{tabular}
    }
    \caption{\label{tab:multi-seg}\textbf{\Multiseg{}.} The task and metrics are described \autoref{sec:background:def:multi-object-segmentation}. We note the detector model used in the column `Detector': `FRCNN' stands for Faster R-CNN~\cite{ren2015fasterrcnn}, `MRCNN' for Mask R-CNN~\cite{he2017mask}, `C-MRCNN' for Cascade Mask R-CNN~\cite{cai2018cascade} and SOLO model is described in \cite{wang2020solo}. If the score was not reported in the original paper, we cite in \textcolor{Purple}{purple} the paper which produced the score.
    }
\end{table*} 

\subsubsection{\Multiclassif{}}
\OS{
We now evaluate the ability of the described unsupervised strategies to detect independent objects with the task of \multiclassif{}. We report the state-of-the-art results in \autoref{tab:multi-det} and compare different unsupervised methods which require or not a training step. We reproduce here the scores available in the literature. Note that, unfortunately, not all methods have been compared in the same setup.
}

\OS{
We report results on the datasets COCO 20K \cite{coco2014,vo2020unsup_multi_object_discovery}, PASCAL VOC07 \cite{pascal-voc-2007}, PASCAL VOC12 \cite{pascal-voc-2012} and COCO val2017 \cite{coco2014} using the different metrics described in \autoref{sec:background:def:multi-object-classif}.
Overall CutLer \cite{wang2023cutler} achieves the best results on all datasets, which could be due to its different rounds of \gilles{self}-learning or 
its loss mechanism alleviating noise from missing pseudo-boxes. 
}

\OS{Although these results are encouraging and show the possibility to achieve interesting multi-object detection without any annotation, there is still a gap to close to meet the results of fully-supervised strategies.}

\subsubsection{\Multiseg{}}
\OS{
We now evaluate on the task of \multiseg{} and gather all available results in \autoref{tab:multi-seg}.}

\OS{We observe again that CutLer appears to achieve the best results on all datasets.} \OS{Moreover, the results on the 
challenging UVO video dataset are close to the fully-supervised results. Indeed a SOLOv2 \cite{solov2} trained on COCO in fully-supervised fashion achieves $38.0$ AP$_{50}$ and Mask-RCNN in the same setup achieves $31.0$ as reported in \cite{wang2023cutler}. The results of $22.8$ AP$_{50}$ of CutLer and $14.2$ of Exemplar FreeSOLO are remarkable for a fully \emph{unsupervised} strategy. The dataset features camera shakes, dynamic backgrounds, and motion blur, which might means that learning without any annotation bias (e.g., on `perfect' COCO dataset) could be helpful in such lower quality data.}

\subsection{Discussion and limits}
\label{sec:learning-results}

\OS{We have discussed in this section the opportunity offered by well designed training schemes to improve and regularize the quality of the unsupervised predictions. Indeed, depending on the downstream task, it is possible to adapt `coarse' predictions extracted with little cost from self-supervised features. We also have seen that training an object detector or instance segmentation model enables accurate localization of multiple objects per image, even in more challenging scenarios (e.g., in UVO~\cite{wang2021uvo} dataset).} 

While the results are promising, the results have yet to reach the level of full supervision.
To enhance results, one may wonder how we can:
\begin{itemize}
    \item Improve the features specifically from the task? 
    \item Leverage different modalities in order to obtain more supervision signals?
    \item Provide class information for the detected objects? 
\end{itemize}
We discuss these ideas in \autoref{sec:perspective}. 

\section{Conclusion and discussions}
\label{sec:discussion}

\subsection{Summary}

In this survey, we have reviewed the power of self-supervised \gilles{ViTs}
and the opportunity they provide to perform \emph{object localization} without any manual annotation.
After introducing relevant tasks and metrics in \autoref{sec:background}, we reviewed in \autoref{sec:ssonly} methods that directly use self-supervised representations without any training, and produce coarse masks or boxes. 
Then, we detailed techniques to further refined these masks with self-training strategies in \autoref{sec:training}.

\gilles{We complement our tour of these methods by presenting in \autoref{tab:ssl-features-used} the self-supervised features that they use.}
Strikingly, we remark that DINO \cite{caron2021dino} features are \OS{largely}
used, and almost always outperform other self-supervised alternatives.
Interestingly, some works \cite{shin2022selfmask,li2023paintseg} show that various self-supervised features can be combined to benefit from feature complementarity.

\begin{table}[ht!]
    \centering
    \resizebox{\linewidth}{!}{
    \begin{tabular}{@{}l | p{1.8cm} | p{3.cm} @{}}
    \toprule
        Method  & Self-sup. features used & Other choices explored \\ 
    \midrule
        LOST \cite{simeoni2021lost} & DINO \cite{caron2021dino} &  \\ 
        TokenCut \cite{wang2022tokencut} & DINO \cite{caron2021dino} & \\ 
        DSS \cite{melas2022deepsectralmethod} & DINO \cite{caron2021dino} & MoCov3 \cite{chen2021mocov3}, \rn{DINO \cite{caron2021dino}} \\ 
        SelfMask \cite{shin2022selfmask} & DINO \cite{caron2021dino} \& \rn{MoCov2 \cite{mocov2}} \& \rn{SwAV \cite{caron2020swav}} & \\ 
        FreeSOLO \cite{wang2022freesolo} & \rn{DenseCL \cite{wang2021densecl}} & \rn{SimCLR \cite{chen2020simclr}}, \rn{MoCov2 \cite{mocov2}}, \rn{DINO \cite{caron2021dino}}, \rn{EsViT \cite{li2022esvit}}, \rn{supervised} \\ 
        IMST \cite{lim2022imst} & DINO \cite{caron2021dino} & \\ 
        MaskDistill \cite{gansbeke2022maskdistill} & DINO \cite{caron2021dino} & \\ 
        DeepCut \cite{aflalo2022deepcut} & DINO \cite{caron2021dino} & MoCov3 \cite{chen2021mocov3}, MAE \cite{he2022mae} \\ 
        UMOD \cite{kara2023umod} & \rn{DINO \cite{caron2021dino}} & \\ 
        FOUND \cite{simeoni2023found} & DINO \cite{caron2021dino} & \\ 
        MOVE \cite{bielski2022move} & DINO \cite{caron2021dino} & MAE \cite{he2022mae} \\ 
        CutLER \cite{wang2023cutler} & DINO \cite{caron2021dino} & \\ 
        Ex.-FreeSOLO \cite{ishtiak2023examplar-freesolo} & \rn{DenseCL \cite{wang2021densecl}} & \\ 
        UCOS-DA \cite{zhang2023ucos-da} & DINO \cite{caron2021dino} & \\ 
        WSCUOD \cite{lv2023wscuod} & DINO \cite{caron2021dino} & MAE \cite{he2022mae}, MoCov3 \cite{chen2021mocov3} \\ 
        MOST \cite{rambhatla2023most} & DINO \cite{caron2021dino} & \\
        SEMPART \cite{ravindran2023sempart} & DINO \cite{caron2021dino} & DINO-v2 \cite{oquab2023dinov2} \\
        Box-based ref. \cite{gomel2023boxbasedrefinement} & DINO \cite{caron2021dino} & \\
        UOLwRPS \cite{song2023UOLwRPS} & DINO \cite{caron2021dino} or \rn{MoCov2 \cite{mocov2}} & \rn{DINO \cite{caron2021dino}}, \rn{SimSiam \cite{chen2021exploring}}, \rn{BYOL \cite{grill2020bootstrap}}, \rn{MoCov3 \cite{chen2021mocov3}} \\
        PaintSeg \cite{li2023paintseg} & DINO \cite{caron2021dino} \& Stable-Diff \cite{rombach2022high} & DINO-v2 \cite{oquab2023dinov2} \\ 
    \bottomrule
    \end{tabular}
    }
    \caption{\label{tab:ssl-features-used} \textbf{Self-supervised features in each method.} 
    Most utilize ViT architectures, while some employ CNNs, shown in \textcolor{PineGreen!90!black}{green}, with a ResNet-50 backbone unless specified otherwise.}
\end{table}

\subsection{Limitations}

While largely improving the state-of-the-art, the reviewed \OS{unsupervised} methods still suffer from several limitations.

First, the methods heavily rely on the correlation properties of patch features extracted by ViTs. In presence of similar objects, the corresponding features remains highly correlated even if they belong to different instances making it difficult to separate these objects, especially when they are also spatially close to each other.

Second, the masks extracted from ViT features are, by design, coarse because they are defined at patch level (typically 16x16 for the best results). Recent works have shown the interest of working at a higher resolution, e.g., \cite{simeoni2023found,li2023paintseg,ravindran2023sempart} and further improvement might be achievable by improving these techniques.

\OS{Third, we have seen that training segmentation or detection models allows us to regularize the quality of coarse detection over an entire dataset and overall increase the number of good predictions per image. Several works \cite{wang2023cutler, wang2022freesolo} have investigated how to handle the noise in the pseudo-masks used as ground-truth. One aspect that still remains to be tackled in a well-designed loss is the lack of separation between similar and close objects; often a single mask/box is generated for several objects.}

\subsection{Other Perspectives, Extensions and Future Directions}
\label{sec:perspective}

We conclude this survey by listing 
alternative perspectives addressing the challenge of unsupervised object localization, considering potential extensions, and touching upon future research directions.

\subsubsection{What about classes?}
Unsupervised object discovery methods prioritize object localization over semantic class identification. Here we discuss options for unsupervised class discovery alongside object localization.

\parag{Closed-vocabulary.} In a closed-vocabulary setup, different works, \gilles{e.g., \cite{simeoni2021lost,melas2022deepsectralmethod,kara2023umod},}
have naturally employed a $k$-means clustering approach \gilles{to discover pseudo-classes}. In particular, it is possible to crop the regions of interest (provided by any method in \autoref{sec:ssonly} or \autoref{sec:training}) over all images of a dataset and to produce $k$ clusters. Typically the quality of the produced \gilles{clustering} 
can be evaluated using Hungarian matching \cite{kuhn1955hungarian} \gilles{with the ground truth class clusters}. The quality of the clustering can be also be improved through a learning step \cite{simeoni2021lost, hamilton2022stego}. It is to be noted that there is a whole literature about unsupervised or self-supervised semantic segmentation learning~\cite{hamilton2022stego,ziegler2022leopart,ji2019invariant,cho2021picie}, which although related does not fall directly in the scope of this \gilles{survey}. 
\gilles{Yet the most related method to this line of works is maybe} 
STEGO \cite{hamilton2022stego}, which learns a linear projection layer atop a self-supervised, pre-trained image encoder using a contrastive loss to encourage compact clusters and maintain patch-wise relationships among pre-trained features in image pairs.

\parag{Open-vocabulary.} \gilles{In an open-vocabulary setup, }vision-language contrastive approaches, exemplified by CLIP~\cite{radford2021learning}, have facilitated zero-shot object classification using textual descriptions.
Therefore, another option for extending the class-agnostic framework discussed in this survey to a class-aware one involves feeding 
the discovered objects from the methods in this survey into 
CLIP\gilles{-like} model along with relevant textual descriptions of targeted semantic classes~\cite{shin2023namedmask, wysoczańska2023clipdiy}. 
Recent efforts \cite{wang2023diffusion,ma2023diffusionseg} use conditional diffusion models \cite{rombach2022high} which produce high-quality features, and generate class-specific training data \cite{ma2023diffusionseg} or directly discover at inference time the object of interest using as input a text query automatically generated by a VLM, e.g., BLIP \cite{li2022blip}.

\subsubsection{Leveraging different modalities}
The accuracy of localization heavily depends on the quality of self-supervised features. To enhance localization performance, a research avenue is to enrich these features with additional information from different sources and modalities, such as motion features, sound, and depth maps.
These modalities are cheap to acquire, they are easy to align with visual image features, and do not necessitate any manual annotations.

\parag{Motion features.} Some approaches incorporate motion features extracted from videos (e.g., optical flow) to enhance object representation learning during training \cite{karazija2022ppmp,safadoust2023multiobjectdiscovery,zhang2023flowdino,choudhury2022guess_what_moves,bao2023motion_guided_tokens}.
These techniques assume that pixels with similar motion patterns likely belong to the same object. At test time, even when applied to still images, they can effectively identify objects without motion information. 
For example, Zhang et al.\ \cite{zhang2023flowdino} show that the Deep Spectral Method \cite{melas2022deepsectralmethod} applied on FlowDINO features \cite{zhang2023flowdino}, learnt in an unsupervised way on videos, yields better performances on unsupervised object localization tasks than the same method applied to original DINO features \cite{caron2021dino}.

\parag{Self-supervised depth.} Other methods use \emph{self-supervised} depth maps to improve object localization \cite{safadoust2023multiobjectdiscovery,hoyer2021threeways,hoyer2023threeways_ijcv}.
This is based on the simple observation that depth discontinuities often coincide with object borders \cite{vandenhende2022dense_prediction_survey,chen2019semantic_synthetic}. 

\parag{Lidar.} Alternatively, 
\OS{rich} lidars \OS{(often available in driving datasets)} 
provide \gilles{3D information about the structure of the scene in form of point clouds,}
offering \gilles{additional} resources for unsupervised object localization and segmentation in images~\cite{vobecky2022drive,tian2021unsupervised}.

\parag{Sound.} can also offer valuable information for object localization. It has actually been shown a supervisory signal can be extracted from audio-visual data by utilizing the audio component to guide object localization~\cite{triantafyllos2020self,chen2021localizing}. In essence, these approaches \gilles{are based on the identification of }
the source of sounds within images~\cite{arandjelovic2018objects,aytar2016soundnet,kidron2005pixels,owens2016visually}.

\subsubsection{Going beyond object-centric datasets}
Self-supervised methods for feature extraction are typically trained on datasets that predominantly focus on objects at the center of images. This training approach introduces a ``collection bias" where the characteristics of real-world images and these collected images may not align well.
Despite this challenge, recent developments are promising. For instance, FOUND \cite{simeoni2023found} shows that it is possible to detect objects that are not typically found in the training data, such as dinosaurs or UFOs that are not part of ImageNet. It can also locate objects that are partially cropped or located close to the image's boundaries.
Furthermore, Zhang et al. \cite{zhang2023ucos-da} evaluate various unsupervised object localization techniques, including FOUND \cite{simeoni2023found} and TokenCut \cite{wang2022tokencut}, on datasets featuring camouflaged objects. It is encouraging to see that these methods perform reasonably well in localizing such objects.

To address situations where images significantly deviate from the training distribution, such as due to defocus blur, heavy obstruction, or cluttered scenes, the VizWiz-Classification dataset \cite{bafghi2023vizwiz-classification} presents a valuable resource. This dataset consists of photos taken by visually impaired individuals and is less likely to exhibit the previously mentioned collection bias.
Another avenue of research involves training or fine-tuning features on real-world scene-centric images \cite{gupta2019lvis}. For instance, WSCUOD \cite{lv2023wscuod} and ORL \cite{xie2021objectrepresentation_sceneimage} mitigate the object-centric bias by training on scene images with complex backgrounds, as they discovered that DINO features are sensitive to intricate backgrounds.

\subsubsection{Learning object-centric features}
While self-supervised learning methods like DINO \cite{caron2021dino} and MAE \cite{he2022mae} focus on learning general representations suitable for any downstream tasks, there is a subset of methods dedicated to learning object-aware features. 
One approach is the use of `slot' methods \cite{locatello2020slotattention}, which employ a structured latent space to promote the learning of object-centric features.
Recent examples include SlotCon \cite{wen2022slotcon}, Odin \cite{henaff2022odin}, DIVA \cite{lao2023diva}, and DINOSAUR \cite{seitzer2022dinosaur}. It is worth noting that these methods often emphasize semantic, i.e., class-based, slots rather than individual instances.
Besides, another technique involves models like VQ-VAE \cite{vandenoord2017vq-vae} or VQ-GAN \cite{esser2021vq-gan} which learn a discrete, structured representation of images. This `quantized' representation resides in a low-dimensional space with meaningful semantics and less variability compared to the original color space. These features in the embedding space offer a promising opportunity for improving localization tasks. In particular, we note a recent work \cite{bao2023motion_guided_tokens} that builds on these quantized representations, jointly with motion cues, to disentangle objects from background. 
Leveraging diffusion models, recent methods \cite{jiang2023object, wu2024slotdiffusion} have investigated the benefits of decoding slots using a latent diffusion model and shown higher-quality outputs.

\subsubsection{Other localization applications}
Self-supervised features have greatly improved object localization in images. However, we see potential in extending their use beyond 2D visuals. For instance, AutoRecon \cite{wang2023autorecon} advances unsupervised 3D object localization in object-centric videos. It begins by coarsely segmenting the salient foreground object from a Structure-for-Motion (SfM) point cloud, using 2D DINO features at the point level. Then, it refines foreground masks consistently across multiple views through neural scene representation.
Another example is the task of unsupervised object localization in videos. It can benefit from high-quality self-supervised features and draw inspiration from methods designed for still images.
VideoCutLer \cite{wang2023videocutler} is a notable step in this direction. It initially generates pseudo-masks in images using MaskCut \cite{wang2023cutler} and TokenCut \cite{wang2022tokencut}. Videos of pseudo mask trajectories are then produced and used to train a video instance segmentation model.

\subsubsection{Conclusion}
In summary, this survey has provided a comprehensive review of the literature on methods for unsupervised object localization in the era of self-supervised ViTs. We have discussed various approaches to exploit self-supervised features for unsupervised object localization, demonstrating promising results across different evaluation setups. By highlighting the potential and opportunities presented by this task, we hope to inspire further research in this direction.

\bmhead{Acknowledgments}
This work was partially supported by the ANR MultiTrans project (ANR-21-CE23-0032). \OS{We would like to thank fellow researcher Étienne Meunier for the rich conversations around self-supervised features.}

\section*{Declarations}

This survey was funded by Valeo and no other funding was received to assist with the preparation of this manuscript. The authors have no relevant financial or non-financial interests to disclose.

{\small
\bibliographystyle{abbrvnat}
\bibliography{main}
}

\end{document}